%% file: main.tex
\documentclass[10pt,twocolumn,letterpaper]{article}

\usepackage{cvpr}              

\input{preamble}

\definecolor{cvprblue}{rgb}{0.21,0.49,0.74}
\usepackage[pagebackref,breaklinks,colorlinks,allcolors=cvprblue]{hyperref}

\usepackage[utf8]{inputenc} 
\usepackage[T1]{fontenc}    
\usepackage{url}            
\usepackage{booktabs}       
\usepackage{amsfonts}       
\usepackage{nicefrac}       
\usepackage{microtype}      
\usepackage{xcolor}         
\usepackage{mathtools} 

\usepackage{graphicx}
\usepackage{amsmath} 
\usepackage{multirow}
\usepackage{colortbl}  
\usepackage{array}   
\usepackage{booktabs, makecell, multirow}
\usepackage{tabularx}

\usepackage{colortbl}
\newcommand{\best}[1]{\colorbox{pink}{#1}}
\newcommand{\second}[1]{\colorbox{yellow}{#1}}
\definecolor{grassgreen}{rgb}{0.4, 0.8, 0.1}

\def\paperID{18180} 
\def\confName{CVPR}
\def\confYear{2026}

\title{ReFlow: Self-correction Motion Learning for Dynamic Scene Reconstruction}

\author{
    Yanzhe Liang$^{1}$ \hspace{6pt} 
    Ruijie Zhu$^{1}$ \hspace{6pt} 
    Hanzhi Chang$^{1}$ \hspace{6pt} 
    Zhuoyuan Li$^{1}$ \hspace{6pt} 
    Jiahao Lu$^{1}$ \hspace{6pt} 
    Tianzhu Zhang$^{1,2}$\thanks{Corresponding author.} \\[5pt]
    $^{1}$University of Science and Technology of China \\
    $^{2}$National Key Laboratory of Deep Space Exploration, Deep Space Exploration Laboratory \\
}

\begin{document}
\maketitle
\input{sec/0_abstract}

\input{sec/1_intro}
\input{sec/2_related_work}
\input{sec/3_method}

\input{sec/4_exp}
\input{sec/5_conclusion}

\section*{Acknowledgements}
This work was supported by the National Natural Science Foundation of China (No. U25A20536), the Youth Innovation Promotion Association of CAS, and the Key Technology Research Project of TW-3 (No. TW3003).

{
    \small
    \bibliographystyle{ieeenat_fullname}
    \bibliography{ref}
}

\input{sec/X_suppl}

\end{document}

%% file: preamble.tex
\usepackage{adjustbox}



%% file: sec/0_abstract.tex
\begin{abstract}
We present ReFlow, a unified framework for monocular dynamic scene reconstruction that learns 3D motion in a novel self-correction manner from raw video. 
Existing methods often suffer from incomplete scene initialization for dynamic regions, leading to unstable reconstruction and motion estimation, which often resorts to external dense motion guidance such as pre-computed optical flow to further stabilize and constrain the reconstruction of dynamic components. However, this introduces additional complexity and potential error propagation.
To address these issues, ReFlow integrates 
a Complete Canonical Space Construction module for enhanced initialization of both static and dynamic regions, and a Separation-Based Dynamic Scene Modeling module that decouples static and dynamic components for targeted motion supervision.
The core of ReFlow is a novel self-correction flow matching mechanism, consisting of Full Flow Matching to align 3D scene flow with time-varying 2D observations, and Camera Flow Matching to enforce multi-view consistency for static objects. Together, these modules enable robust and accurate dynamic scene reconstruction.
Extensive experiments across diverse scenarios demonstrate that ReFlow achieves superior reconstruction quality and robustness, establishing a novel self-correction paradigm for monocular 4D reconstruction. Project page: \href{https://rosetta-leong.github.io/ReFlow_Page/}{this https URL}

\end{abstract}
\begin{figure}[t]
\centering
\includegraphics[width=0.8\linewidth]{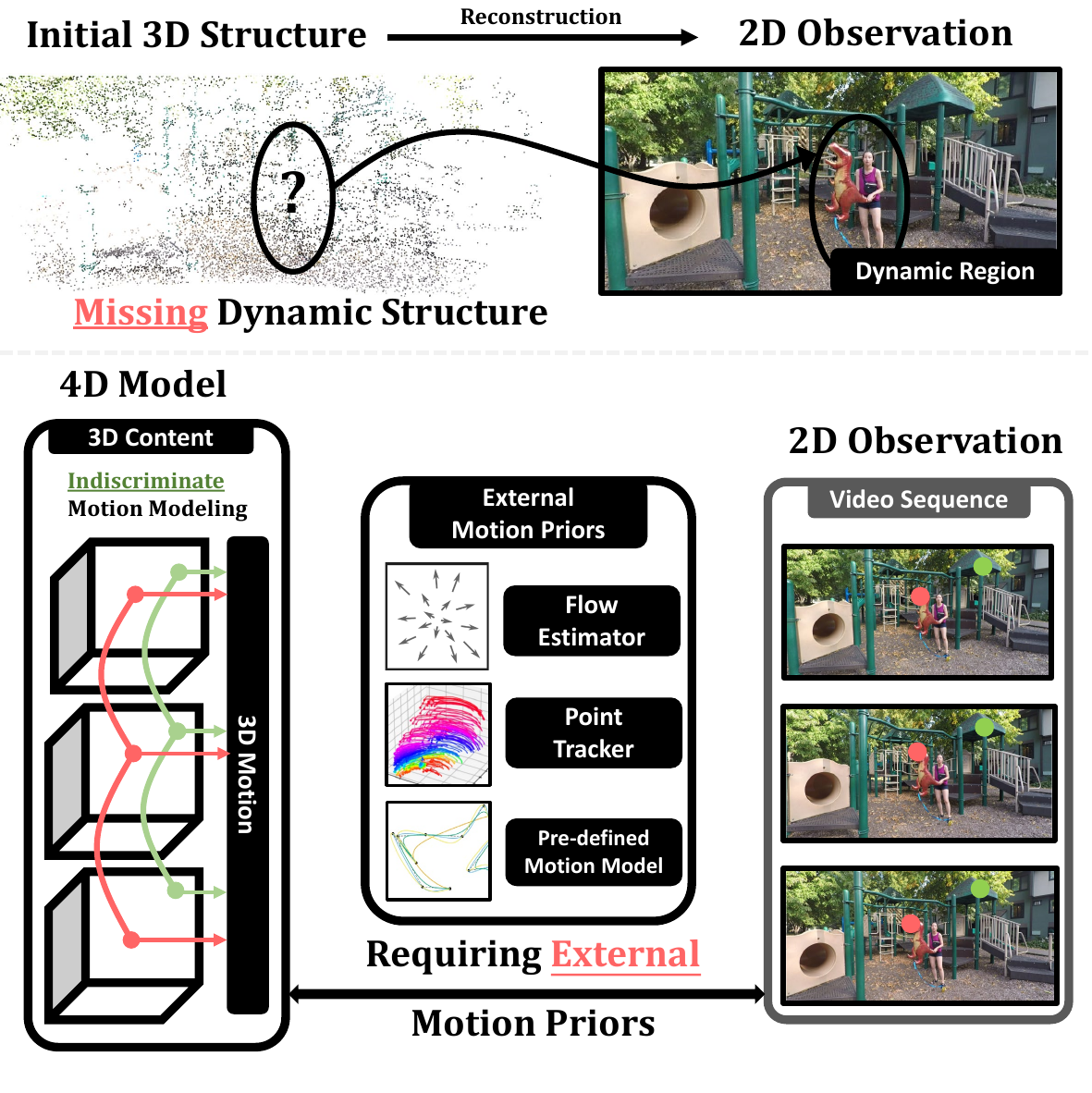}  
\caption{
Typical Challenges in monocular dynamic scene reconstruction. \textbf{Top:} Incomplete initialization for dynamic regions: the initial 3D structure from SfM often misses dynamic components and initializes Gaussians without separating {\textcolor{grassgreen}{\textbf{static points (green)}}} from {\color{pink}\textbf{dynamic points (red)}}, leading to an entangled and incomplete representation. \textbf{Bottom:} To compensate, existing methods frequently resort to external dense motion guidance to constrain and stabilize the reconstruction of dynamic regions.
}
\label{fig:Challenge}
\vspace{-0.5cm} 
\end{figure}

%% file: sec/1_intro.tex
\section{Introduction}
\label{sec:intro}

Monocular dynamic scene reconstruction is a key task in 3D computer vision and graphics, aiming to recover time-varying 3D structure from 2D video sequences. Despite its broad applications in spatial computing~\cite{li2025scenesplat, cheng2026adaptive, cheng2026rethinking, cheng2025i2p, cheng2025bridge}, autonomous driving~\cite{chen2024omnire, chen2026periodic, yan2024street}, and AR/VR~\cite{jiang2024vr}, accurately reconstructing dynamic scenes remains a significant challenge due to the complexity of spatiotemporal modeling and limited observations.

Recent advances in neural rendering~\cite{aliev2020neural, mildenhall2021nerf,barron2021mip, barron2022mip, barron2023zip, mueller2022instant, xu2022point, kerbl20233d} have significantly improved static scene reconstruction, particularly through 3D Gaussian Splatting (3DGS)~\cite{kerbl20233d}. 
Building on this success, several works have extended these techniques to dynamic
scenes~\cite{luiten2023dynamic, yang2023deformable, wu20234d, yang2023real, li2024spacetime}, typically representing scenes with time-varying 3D gaussians to capture temporal changes.
Despite promising results, these approaches still struggle with complex scene structures and motion patterns.
As shown in Fig.~\ref{fig:Challenge}, current approaches typically initialize 3DGS primitives using classical Structure-from-Motion (SfM) methods like COLMAP, which are designed for static scene reconstruction 
but not suitable for dynamic scenes. The initial point cloud produced by COLMAP often lacks dynamic regions and is used to initialize Gaussians without distinguishing static from dynamic content. This incomplete and entangled initialization makes subsequent reconstruction and motion estimation difficult and unstable. As a result, many recent works resort to external dense motion supervision(optical flow~\cite{zhou2024dynpoint, wang2024gflow, sun2025splatter, gao2024gaussianflow, zhu2025motiongs, chen2024freegaussian}, tracking~\cite{wang2024shape, lei2024mosca}) to ensure the accurate reconstruction of scene dynamics.
These motion cues are typically generated by separate pipelines~\cite{xu2022gmflow, huang2022flowformer, xu2023unifying, karaev2024cotracker, karaev2024cotracker3, doersch2023tapir} and are treated as pseudo ground-truth, enforcing hard constraints that force the reconstructed motion to align with potentially flawed external estimates. This introduces additional complexity and potential error propagation into the reconstruction process making the reconstruction highly sensitive to the accuracy of external estimators.
For instance, MotionGS\cite{zhu2025motiongs} decouples optical flow into motion and camera flow, effectively constraining the motion of 3DGS. But it still depends on externally extracted motion flow labels, making its performance tightly coupled with the quality of the flow estimator and prone to failure when the estimator is inaccurate.
This makes us wonder: \textit{\textbf{is it possible to unlock 4D dynamic scenes purely from 2D observations, without external motion guidance?}}

\begin{figure}[t]
\centering
\includegraphics[width=1.0\linewidth]{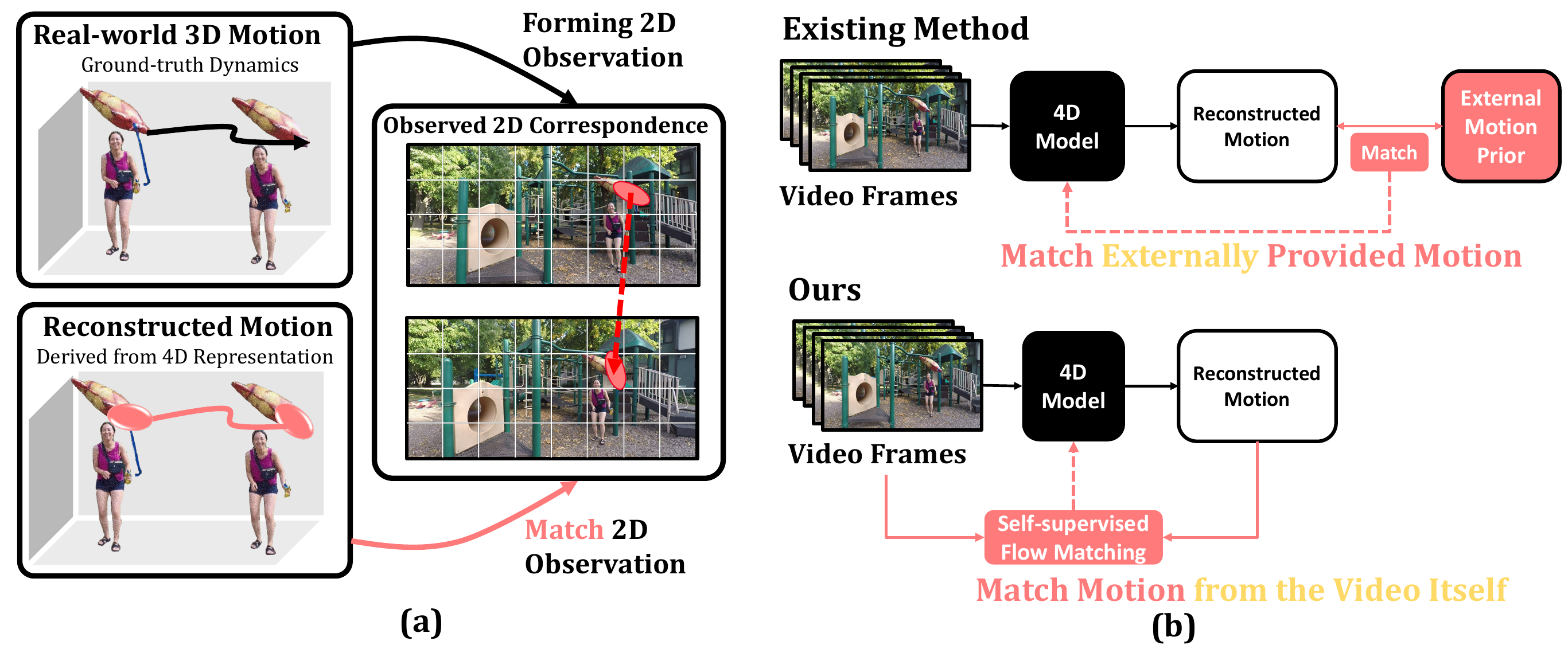} 
\caption{
    \textbf{Motivation of Self-correction Flow Matching.} 
    (a) We start with a simple observation: 2D observations, such as the shifting balloon, are caused by 3D motion. Accurate reconstructed 3D Motion should naturally align with these visible changes.
    (b) Unlike previous methods that use external motion priors to supervise 3D motion,
   we instead uses raw \textbf{video as motion supervision} through a self-correction flow matching mechanism to directly align predicted 3D motion projections with 2D frame differences.
  }
\label{fig:motivation}
\vspace{-0.5cm}
\end{figure}

To answer this question, we revisit the 4D dynamic scenes from both structural and motion perspectives.
Structurally, it is crucial to initialize both static and dynamic regions and to model them with separate representations. Such initialization and disentanglement reduce early-stage ambiguity and yield more stable optimization, while also preserving the potential to apply different motion constraints on different regions.
In terms of motion, we identify a natural 3D-2D motion consistency that can be exploited for self-correction motion constraints, without the need for aligning with external motion priors. 
As shown in Fig.~\ref{fig:motivation}(a), differences between consecutive video frames originate from real-world 3D motion. \textit{\textbf{If 3D motion is accurately reconstructed, it should align with the pixel-wise differences across frames.}} 
This insight leads to our core idea: we directly supervise 3D motion by measuring how well it explains frame-to-frame variations. Specifically, we project 3D motion into a 2D flow field to warp one frame toward the next. The misalignment between the warped frame and the real image provides a natural supervision signal to correct the reconstructed motion, creating a self-correction loop in Fig.~\ref{fig:motivation}(b) where motion is learned and refined by a direct and natural objective of re-rendering future or past frames, rather than imitating a potentially flawed external motion estimations. 

We instantiate these considerations in \textit{\textbf{ReFlow}}, a unified dynamic scene reconstruction framework that integrates three key components.
Firstly, we integrate a \textbf{Complete Canonical Space Construction} module that leverages geometry foundation models~\cite{wang2024dust3r, leroy2024grounding, zhang2024monst3r, lu2024align3r} to initialize both static and dynamic regions in a coarse-to-fine manner.
Secondly, a \textbf{Separation-Based Dynamic Scene Modeling} module independently models static and dynamic parts of the scene, aggregating spatiotemporal information for static regions and enabling flexible motion for dynamic elements. This decoupling also allows us to apply region-specific motion constraints in the next motion learning stage.
On top of these, we propose a natural \textbf{Self-correction Flow Matching} mechanism that learns 3D motion directly from raw video by enforcing consistency between the projected 3D motion and observed 2D frame-to-frame differences. It incorporates two complementary constraints: \emph{Full Flow Matching}, which aligns the joint camera–scene motion with entire image differences, and \emph{Camera Flow Matching}, which enforces consistency between camera-induced motion and the observed changes in static regions separated by the separation module. 
By coupling these constraints within a self-corrective loop, ReFlow derives region-appropriate motion supervision from the video.
Through the integration of initialization, static–dynamic decoupling, and self-correction motion learning, ReFlow presents a unified framework for high-quality monocular dynamic scene reconstruction without dense motion guidance. Our main contributions are summarized as follows:
\begin{enumerate}
    \item We propose ReFlow, a unified framework that addresses incomplete initialization, entanglement, and the overreliance on external motion guidance by combining complete canonical-space initialization, static–dynamic separation, and self-corrective motion learning into a coherent pipeline for high-quality monocular 4D reconstruction.
    \item We formulate and implement a simple motion constraint mechanism for dynamic scene reconstruction in a entirely self-correction way. Instead of relying on external motion priors, our Self-correction Flow Matching directly aligns 3D motion with video frame differences, making video itself as the direct motion supervision. 
    \item Through extensive experiments, we demonstrate ReFlow's superior performance in diverse dynamic scenes, significantly improving reconstruction accuracy and quality, while effectively reducing reliance on external dense motion guidance.
\end{enumerate}

%% file: sec/2_related_work.tex
\section{Related Work}
\label{sec:related_work}
\subsection{Dynamic Scene Reconstruction}
With rapid advances in neural rendering, dynamic scene reconstruction has witnessed significant progress in recent years. Early research primarily extended static NeRF to dynamic scenes along two technical routes: one approach~\cite{cao2023hexplane,pumarola2021d,fridovich2023k} constructs a 3D canonical space to represent scene content, complemented by independent motion modules for temporal modeling; the other~\cite{xian2021space, li2023dynibar} directly builds complete 4D spatiotemporal representations, modeling time-varying spatial structures in a unified framework. Despite establishing foundations, NeRF-based methods faced limitations from their implicit nature, including computational complexity and slow inference.
3D Gaussian Splatting's success introduced a more efficient paradigm for dynamic scenes. Recent approaches adopt 3DGS representations for canonical space with temporal deformation, modeled either implicitly through neural networks~\cite{yang2023deformable, huang2024sc, liang2023gaufre, jiang2024timeformer} and feature planes~\cite{wu20234d}, or explicitly via velocity fields~\cite{chen2023periodic} and various trajectory models~\cite{lin2024gaussian,hu2024learnable, bond2025gaussianvideo, li2024spacetime, katsumata2024compact}, .
In parallel, alternative approaches~\cite{yang2023real} directly model time-varying 4D Gaussian distributions, offering a different perspective on the problem. 
Building on recent advances in dynamic Gaussian Splatting, we take a closer look at how the different roles of dynamic and static components, as well as complete canonical space construction strategies, can provide a more reliable foundation for learning motion constraints in a self-correction manner.

\subsection{Motion Regularization in 4D Scene}
\begin{figure*}[t]
\centering
\includegraphics[width=1.0\linewidth]{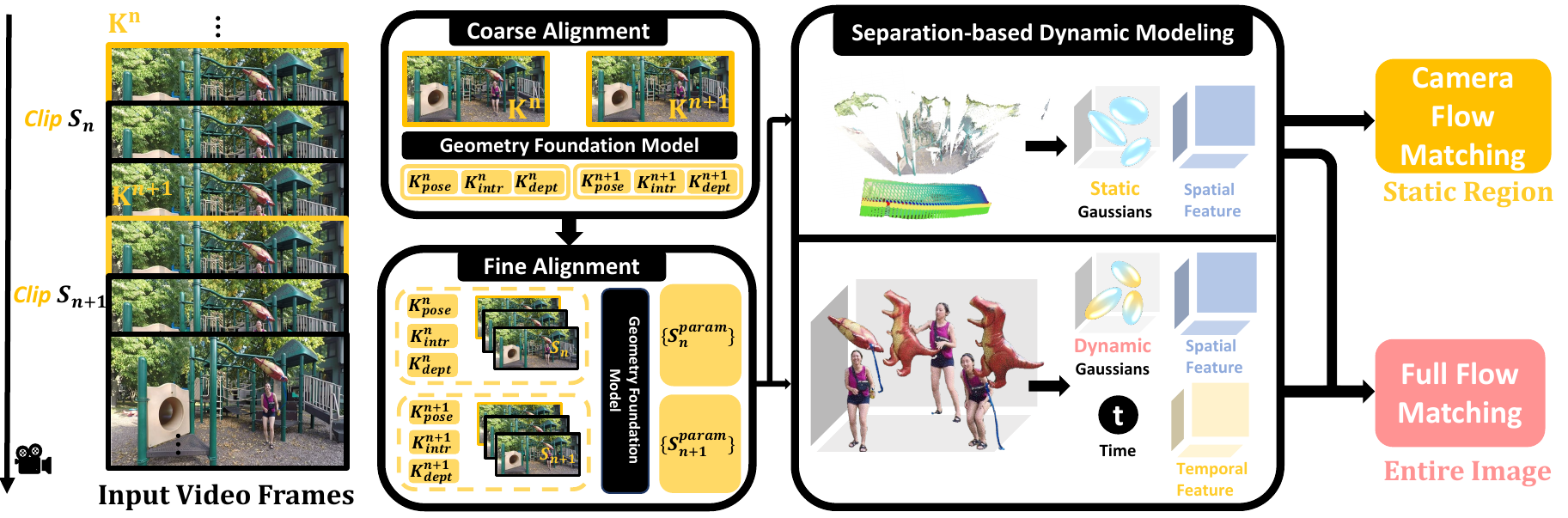}  
\caption{
\textbf{Overview of \textit{ReFlow}.} 
We start by constructing a complete canonical space(Sec.~\ref{sec:canonical}), which includes both static and dynamic components, ensuring a reliable 3D scene initialization.
Next, we disentangle these elements using spatial and spatiotemporal feature planes(Sec.~\ref{sec:modeling}), providing a structured representation that separately handles static and dynamic regions.
This preparation allows us to introduce targeted motion constraints(Sec.~\ref{sec:flowmatching}): Full Flow supervises motion across the entire scene, while Camera Flow enforces consistency in static regions, enabling the self-correction learning mechanism for accurate 3D motion reconstruction.
}
\label{fig:pipeline}
\end{figure*}

Motion constraint mechanisms play a decisive role in dynamic scene reconstruction, directly affecting the temporal consistency and physical plausibility of reconstruction results. Existing methods primarily constrain scene dynamics by introducing external 2D priors, broadly categorized into optical flow-based and trajectory-based approaches.
Optical flow-based methods~\cite{li2023dynibar, zhou2024dynpoint, wang2024gflow, gao2024gaussianflow, zhu2025motiongs} derive inherent optical flow from scene representations and establish correspondences with pre-computed 2D optical flow priors~\cite{xu2022gmflow,huang2022flowformer, Ranftl2022}, constraining 3D motion through photometric consistency loss. While capturing fine-grained motion, they depend heavily on flow estimation quality and demonstrate sensitivity to rapid movements and occlusions.
Trajectory-based methods~\cite{lei2024mosca, wang2024shape} lift 2D point tracking~\cite{karaev2024cotracker, karaev2024cotracker3} to 3D space to guide Gaussian trajectories. While these methods show advantages in handling complex deformations, they typically require stable estimated point trajectories and perform poorly in textureless or rapidly moving regions. Unlike these approaches, our method employs a self-correction strategy extracting motion information directly from raw inputs, ensuring reconstruction quality while simplifying the process. 

%% file: sec/3_method.tex
\section{Methodology}
\label{sec:method}
\subsection{Overview}
Given a monocular video sequence $\mathcal{V} = \{I_i\}_{i=1}^{N}$ with extrinsics $\mathbf{T} = \{\mathbf{T}_i \in \mathbb{R}^{3\times4}\}_{i=1}^{N}$ and intrinsics $\mathbf{K}$, our goal is to reconstruct a dynamic 3D scene and learn its motion in a fully self-correcting manner.
We particularly target two practical challenges in monocular 4D reconstruction: (i) the lack of reliable initialization for dynamic regions and the need for decoupled representations of static and dynamic content, and (ii) the strong dependence on external dense motion guidance.
To tackle these issues, we propose ReFlow, as illustrated in Fig.~\ref{fig:pipeline}.
It consists of three key modules that together enable proper scene initialization and decoupling, along with a novel self-correction of motion directly from video. The details of each module are presented in the following sections.

\begin{figure*}[t]
\centering
\includegraphics[width=1.0\linewidth]{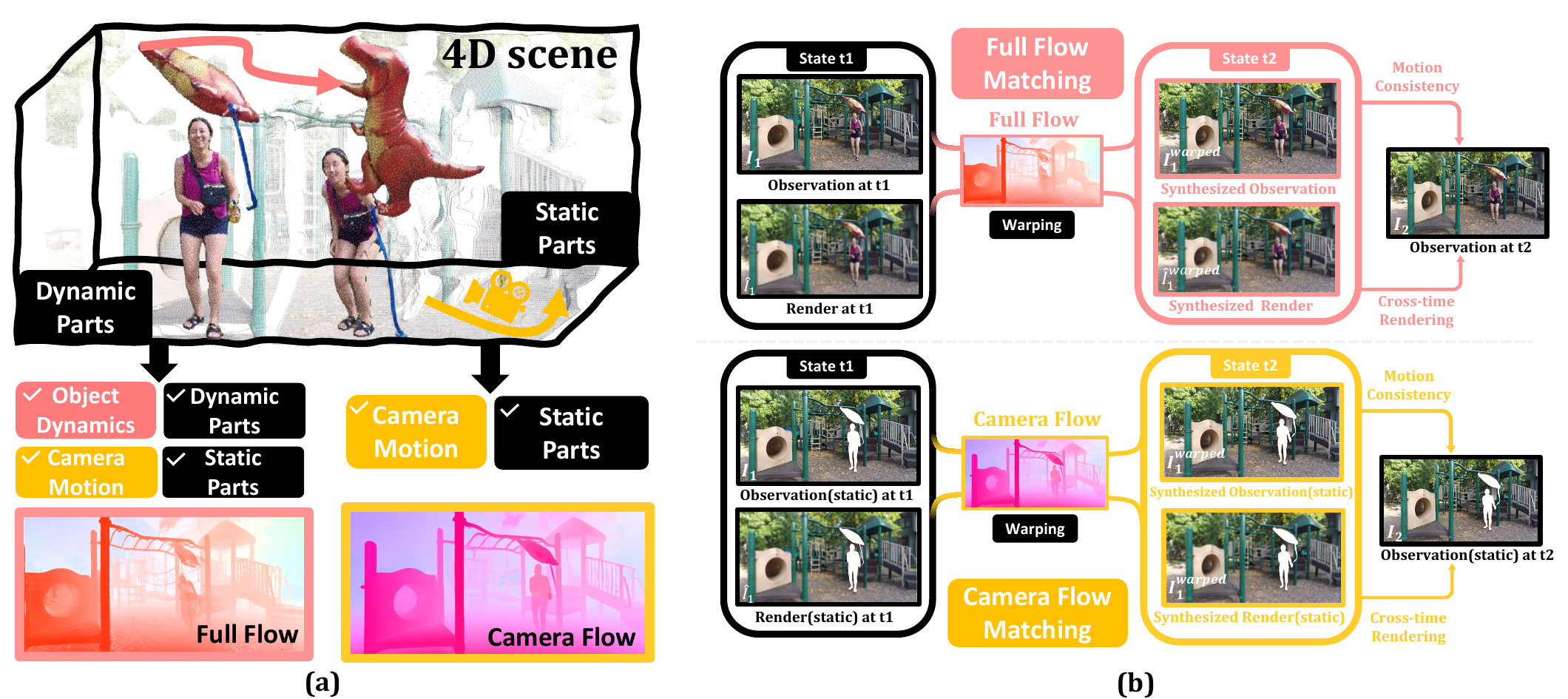}  
\caption{
\textbf{Self-correction flow matching mechanism.}
(a) \textbf{Different Motion and Flow in the 4D Scene.} 
Static areas move only due to camera motion (camera flow), while dynamic areas involve both camera and object motion (full flow). Accurate motion learning requires region-specific flow supervision.
(b) \textbf{Self-correction flow matching.} We apply full flow to warp the entire image from state $t_{1}$ to state $t_{2}$ and compare with the real observation, validating overall motion. Camera flow is used similarly but only on static regions, ensuring their stability. Together, these provide a complementary self-correction signal for 3D motion learning.
}
\label{fig:flow}
\end{figure*}

\subsection{Dynamic Scene Initialization and Modeling}
\subsubsection{Complete Canonical Space Construction}
\label{sec:canonical}
To address the challenge of obtaining a proper initialization for dynamic scenes, we first build a complete canonical space that consistently captures both static and dynamic regions across the entire video.
We leverage a geometry foundation model~\cite{zhang2024monst3r} to regress per-pixel 3D coordinates from image pairs and enforce geometric consistency over a global connectivity graph to obtain a unified point cloud. Directly applying this to videos faces two challenges: (1) memory cost grows quadratically ($O(N^2)$) with frame number; (2) limited co-visibility across distant views hinders alignment. We address these via a hierarchical coarse-to-fine strategy.

\noindent \textbf{Coarse-to-fine Alignment.}  
Given a video sequence $\{I_i\}_{i=1}^N$, we divide it into $K$ non-overlapping clips $\{S_k\}_{k=1}^K$ and select the first frame $I_{s_k}$ of each clip as a keyframe. We build a smaller connectivity graph $G_K$ among keyframes to establish a coarse globally consistent structure. For each keyframe pair $(I_a, I_b)$, the geometry foundation model regresses 3D coordinates $\mathbf{X}_a^a, \mathbf{X}_a^b$ in $a$'s coordinate system, alongside confidence maps $\mathbf{C}_a^a, \mathbf{C}_a^b$. Global alignment is obtained by optimizing keyframe camera poses $K_{\text{pose}}$, intrinsics $K_{\text{intr}}$, and depth maps $K_{\text{depth}}$ using the 3D geometry consistency constaint $\mathcal{L}_{\text{align}}$ from~\cite{zhang2024monst3r}:
\[
\min_{K_{\text{pose}}, K_{\text{intr}}, K_{\text{depth}}} \sum_{(a,b) \in E_K} \mathcal{L}_{\text{align}}(\mathbf{X}_a^a, \mathbf{X}_a^b, \mathbf{C}_a^a, \mathbf{C}_a^b)
\]
We then recover geometry for all frames via local alignment within each clip $S_k$, using a similar loss initialized from keyframe alignment. This hierarchical design enforces both local consistency between temporally close frames and global consistency across the entire sequence. Further details are provided in Appendix~\ref{init_explanation}.

\noindent \textbf{Static-Dynamic Disentanglement.}
We separate static and dynamic elements using dynamic masks $M_i^{\text{dyn}}$ from the geometry foundation model for each frame $i$ in clip $S_k$. Each frame is back-projected to 3D using depth maps $S_k^{\text{depth}}$, camera intrinsics $S_k^{\text{intr}}$, and poses $S_k^{\text{pose}}$. Aggregating all clips, we obtain separate point clouds for static content $\mathbf{P}^{3D,\text{stat}} = \{\mathbf{P}_i^{3D,\text{stat}}\}_{i=1}^N$ and dynamic regions $\mathbf{P}^{3D,\text{dyn}} = \{\mathbf{P}_i^{3D,\text{dyn}}\}_{i=1}^N$, which 
enable subsequent region-specific modeling of static and
dynamic components.

\subsubsection{Separation-Based Dynamic Scene Modeling}
\label{sec:modeling}
With the aligned point clouds $\mathbf{P}^{3D,\text{stat}}$ and $\mathbf{P}^{3D,\text{dyn}}$, we adapt separate representations for static and dynamic components, enabling motion-aware modeling and preparing the representation for motion-specific constraints.

\noindent\textbf{Separate Initialization.} To initialize our Gaussian models, we process the point cloud sequences into unified representations. 
For dynamic objects, we select a reference frame with maximal mask coverage and color diversity to comprehensively represent $\mathbf{P}^{3D,\text{dyn}}$. For static regions, point clouds from current and keyframes are fused into $P^{3D,\text{stat}}_{\text{fusion}}$ to increase viewpoint coverage.

\noindent\textbf{Modeling Static and Dynamic Components.}  
Static components use tri-plane spatial features $F_s = \{F^{xy}, F^{xz}, F^{yz}\}$ encoding position-dependent information. Dynamic components add temporal planes $F_t = \{F^{xt}, F^{yt}, F^{zt}\}$ for motion modeling:
\[
G_s: \{\mu_s, s_s, q_s, \sigma_s, c_s\} = D_s(x,y,z; F_s)
\]
\[
G_d: \{\mu_d(t), s_d, q_d(t), \sigma_d, c_d\} = D_d(x,y,z,t; F_s, F_t)
\]
where $D_s$, $D_d$ decode static and dynamic Gaussian parameters respectively. We provide detailed explanation in Appendix~\ref{model_explannation}. This representation cleanly separates static structures from time-varying dynamic parts and provides exactly the representation needed for the motion-specific constraints introduced next in our self-correction flow matching module.

\subsection{Self-correction Flow Matching}
With coherent initialization and static–dynamic separation, the remaining goal is to learn 3D motion consistent with the actual temporal evolution of the video. Instead of relying on external optical flow or tracking, we aim to explore the capability of a motion learning mechanism derived purely from the raw video.

\label{sec:flowmatching}
\subsubsection{Formulation}
\label{sec:flow_formulation}
\noindent\textbf{Formulate 3D–2D Alignment as Warping.}  
Given consecutive frames \(I_1\) and \(I_2\), an accurate 3D motion should explain the observed image changes. If the estimated motion reflects real-world dynamics, it should induce a 2D flow that warps \(I_1\) to closely match \(I_2\). This 3D--2D alignment can be realized as a simple pixel-wise warping: projecting 3D motion into 2D flow and warping one frame toward the other, then aligning the warped image with the target frame.
We name this process \textbf{self-correction flow matching}, where the reconstructed 3D motion aligns with the observed visual changes between input video frames rather than external motion estimation, using the video itself as a self-correction signal for learning 3D motion.

\noindent\textbf{Constructing Flows for Diverse Motion Types.}  
As our self-correction matching relies on deriving 2D flow, and following prior work~\cite{gao2024gaussianflow,zhu2025motiongs, chen2024freegaussian}, we distinguish two types of motion—object motion and camera ego-motion and define two corresponding flow fields to separately capture their effects(see Fig.~\ref{fig:flow}(a)).
\textit{Full Flow} represents the total image-space displacement caused by both object motion and camera movement, corresponding to changes across the entire frame. Given the scene at frames $t_1$ and $t_2$, represented by 3D Gaussians $G_{t_1}, G_{t_2}$, and the corresponding camera parameters $\mathbf{P}_1$, $\mathbf{P}_2$, 
we synthesize Full Flow as
\begin{equation}
\mathbf{F}_{full} = \text{FlowRender}(G_{t_1}, G_{t_2}, \mathbf{P}_1, \mathbf{P}_2),
\end{equation}
where $\mathbf{P}_i = (\mathbf{K}_i, \mathbf{T}_i)$ denotes the intrinsic and extrinsic matrices $\mathbf{K}_i$ and $\mathbf{T}_i$ respectively.
\textit{Camera Flow} isolates the 2D motion induced solely by camera movement, assuming the scene is static. As static regions should remain geometrically stable, Camera Flow provides an unambiguous and clean supervision signal in static regions where observed changes arise only from viewpoint shifts, ensuring structural consistency and avoiding false motion supervision or dynamic overfitting. 
We compute it by warping the static regions of the scene at $t_1$ according to the camera's pose change:
\begin{equation}
\mathbf{F}_{cam} = \text{CamFlowRender}(G_{t_1}, \mathbf{P}_1, \mathbf{P}_2),
\end{equation}
Implementation details for both flow renderings are provided in Appendix~\ref{sec:supp_flow_render_details}. 
In the following sections, we design separate matching mechanisms for each flow type to impose region-specific constraints.

\subsubsection{Full Flow Matching}
Specifically, we define two constraints by warping the input and rendered images, offering supervision from different perspectives.

\noindent\textbf{Motion Consistency Constraint.}
 We warp the image $I_1$ from time $t_1$ to $t_2$ using the synthesized flow field $\mathbf{F}$, producing $I_1^{warped}$. Ideally, if the motion modeling is accurate, $I_1^{warped}$ should be consistent with the real image $I_2$. Our motion consistency loss ${L}_{mc}$ verifies if the derived 2D motion accurately reflects real-world visual changes:
\begin{equation}
\mathcal{L}_{mc} = \mathcal{L}_{photo}(I_1^{warped}, I_2)
\end{equation}
where $\mathcal{L}_{photo}$ combines L1 distance and structural similarity metrics \cite{kerbl20233d}. This warping-based supervision provides an explicit correction signal: whenever the 3D motion is inaccurate, the warped frame will deviate from the observation, pushing the model toward the correct motion field.

\noindent\textbf{Cross-Time Rendering Constraint.}
We further warp the rendered image $\hat{I}_1$ at time $t_1$ through the same flow field, creating a cross-time rendering constraint ${L}_{cr}$:
\begin{equation}
\mathcal{L}_{cr} = \mathcal{L}_{photo}(\hat{I}_1^{warped}, {I_2})
\end{equation}
This enables predicting scene appearance at $t_2$ from its state at $t_1$, reinforcing temporal consistency.
The complete Full Flow Matching objective combines both constraints:
\begin{equation}
\mathcal{L}_{fullflow} = \lambda_{mc}\mathcal{L}_{mc} + \lambda_{cr}\mathcal{L}_{cr}
\end{equation}

\begin{table*}[!t]
\centering
\caption{Quantitative comparison on Nvidia Monocular dataset. We report PSNR/SSIM/LPIPS per scene; the last block shows the mean across all available scenes (including dynamicFace when available).\label{tab:nvidia-monocular}}
\small
\setlength{\tabcolsep}{1.5pt} 
\resizebox{0.8\linewidth}{!}{    
\begin{tabular}{l|ccc|ccc|ccc|ccc}
\toprule
\multirow{2}{*}{Method} & \multicolumn{3}{c|}{Balloon1} & \multicolumn{3}{c|}{Balloon2} & \multicolumn{3}{c|}{Jumping} & \multicolumn{3}{c}{Playground} \\
\cmidrule{2-13}
 & PSNR$\uparrow$ & SSIM$\uparrow$ & LPIPS$\downarrow$ & PSNR$\uparrow$ & SSIM$\uparrow$ & LPIPS$\downarrow$ & PSNR$\uparrow$ & SSIM$\uparrow$ & LPIPS$\downarrow$ & PSNR$\uparrow$ & SSIM$\uparrow$ & LPIPS$\downarrow$ \\
\midrule
3DGS~\cite{kerbl20233d} & 17.71 & 0.407 & 0.473 & 20.62 & 0.500 & 0.431 & 14.45 & 0.394 & 0.385 & 15.75 & 0.341 & 0.493 \\
STGS~\cite{li2024spacetime}  & 20.36 & 0.746 & 0.196 & 23.12 & 0.881 & 0.124 & 20.82 & 0.631 & 0.187 & 19.23 & 0.842 & 0.151 \\
Ex4DGS~\cite{lee2024fully} & 14.69 & 0.351 & 0.503 & 16.29 & 0.471 & 0.457 & 18.93 & 0.661 & 0.321 & 14.16 & 0.442 & 0.437 \\
Efficient-D3DGS~\cite{katsumata2024compact} & 23.32 & 0.864 & 0.140 & 26.20 & 0.906 & \second{0.099} & 21.54 & 0.784 & 0.253 & \second{24.08} & \second{0.892} & 0.098 \\
Deformable-3DGS~\cite{yang2023deformable} & 26.12 & \second{0.891} & \second{0.122} & \best{26.48} & \best{0.910} & 0.100 & \second{23.21} & 0.837 & 0.199 & 23.96 & \best{0.901} & \second{0.088} \\
4DGS~\cite{wu20234d} & 25.46 & 0.856 & 0.198 & 27.12 & 0.842 & 0.151 & 22.43 & 0.842 & 0.264 & 22.17 & 0.743 & 0.215 \\
MoDec-GS~\cite{kwak2025modec} & \second{26.33} & 0.883 & 0.174 & \second{27.15} & \second{0.909} & 0.102 & 23.12 & \best{0.858} & \second{0.198} & 23.33 & 0.816 & 0.150 \\
Ours & \best{27.65} & \best{0.903} & \best{0.092} & \best{29.01} & 0.908 & \best{0.098} & \best{24.85} & \second{0.855} & \best{0.151} & \best{25.58} & 0.889 & \best{0.086} \\
\midrule
\multirow{2}{*}{Method} & \multicolumn{3}{c|}{Skating} & \multicolumn{3}{c|}{Truck} & \multicolumn{3}{c|}{Umbrella} & \multicolumn{3}{c}{Mean} \\
\cmidrule{2-13}
& PSNR$\uparrow$ & SSIM$\uparrow$ & LPIPS$\downarrow$ & PSNR$\uparrow$ & SSIM$\uparrow$ & LPIPS$\downarrow$ & PSNR$\uparrow$ & SSIM$\uparrow$ & LPIPS$\downarrow$ & PSNR$\uparrow$ & SSIM$\uparrow$ & LPIPS$\downarrow$ \\
\midrule
3DGS~\cite{kerbl20233d} & 21.20 & 0.628 & 0.508 & 13.81 & 0.413 & 0.627 & 20.05 & 0.410 & 0.530 & 17.69 & 0.451 & 0.497 \\
STGS~\cite{li2024spacetime} & 24.80 & 0.910 & 0.109 & 25.01 & 0.868 & 0.103 & 21.88 & 0.770 & 0.195 & 22.17 & 0.807 & 0.152 \\
Ex4DGS~\cite{lee2024fully} & 21.92 & 0.807 & 0.233 & 19.04 & 0.666 & 0.308 & 19.03 & 0.593 & 0.340 & 17.72 & 0.570 & 0.371 \\
Efficient-D3DGS~\cite{katsumata2024compact} & 25.44 & 0.892 & 0.161 & 24.33 & 0.829 & 0.232 & 23.95 & 0.805 & \second{0.183} & 24.92 & 0.868 & 0.150 \\
Deformable-3DGS~\cite{yang2023deformable} & 26.53 & 0.929 & \second{0.126} & 24.16 & 0.806 & 0.192 & \second{26.18} & \best{0.861} & 0.187 & 25.86 & \best{0.888} & \second{0.132} \\
4DGS~\cite{wu20234d} & 28.92 & 0.931 & 0.196 & 28.25 & 0.888 & 0.235 & 24.80 & 0.714 & 0.297 & 25.81 & 0.844 & 0.210 \\
MoDec-GS~\cite{kwak2025modec}  & \second{29.29} & \second{0.942} & 0.156 & \second{29.19} & \second{0.910} & \second{0.185} & 25.02 & 0.760 & 0.224 & \second{26.63} & 0.879 & 0.160 \\
Ours & \best{30.17} & \best{0.946} & \best{0.076} & \best{30.18} & \best{0.917} & \best{0.115} & \best{27.21} & \second{0.831} & \best{0.170} & \best{28.20} & \second{0.903} & \best{0.103} \\

\bottomrule
\end{tabular}}
\end{table*}

\subsubsection{Camera Flow Matching} 
Camera Flow Matching refers to a self-correction constraint applied to static regions where image changes result solely from camera motion. By utilizing Camera Flow $\mathbf{F}_{cam}$, which contains only camera-induced motion, we enforce that static areas strictly follow this flow to ensure structural stability and prevent false motion.
Following the Full Flow Matching principle, we define motion consistency and cross-time rendering constraints by warping input and rendered images with $\mathbf{F}_{cam}$, applied exclusively to static regions.
\begin{equation}
\mathcal{L}_{mc}^{cam} = \mathcal{L}_{photo}(I_1^{static,warped}, I_2^{static})
\end{equation}
\begin{equation}
\mathcal{L}_{cr}^{cam} = \mathcal{L}_{photo}(\hat{I}_1^{static,warped}, I_2^{static})
\end{equation}
\begin{equation}
\mathcal{L}_{camflow} = \lambda_{mc}^{cam}\mathcal{L}_{mc}^{cam} + \lambda_{cr}^{cam}\mathcal{L}_{cr}^{cam}
\end{equation}
This mechanism ensures static regions maintain consistent 3D structure throughout the sequence and eliminates false motion caused by lighting or view-change artifacts.
Together, these two flow constraints constitute a complete self-correction loop that enables reliable and dense-guidance-free 3D motion learning.

\subsection{Optimization}
We optimize frame pairs \((I_1, I_2)\) 
with supervision solely from the video frames by the loss:
\begin{equation}
\begin{split}
\mathcal{L} = &\ \mathcal{L}_{baseline} + \lambda_{ff} \mathcal{L}_{fullflow}(I_1, I_2) \\
& + \lambda_{cf} \mathcal{L}_{camflow}(I_1^{static}, I_2^{static})
\end{split}
\end{equation}
where $\mathcal{L}_{baseline}$ denotes the baseline loss from the 4DGS framework~\cite{wu20234d}, 
which consists of a photometric rendering loss and a feature plane regularization term, 
and $\lambda_{ff}$, $\lambda_{cf}$ weight the Full Flow Matching and Camera Flow Matching terms respectively.

%% file: sec/4_exp.tex
\begin{figure*}[t]
    \vspace{1em}
    \centering
    \includegraphics[width=1.0\textwidth]{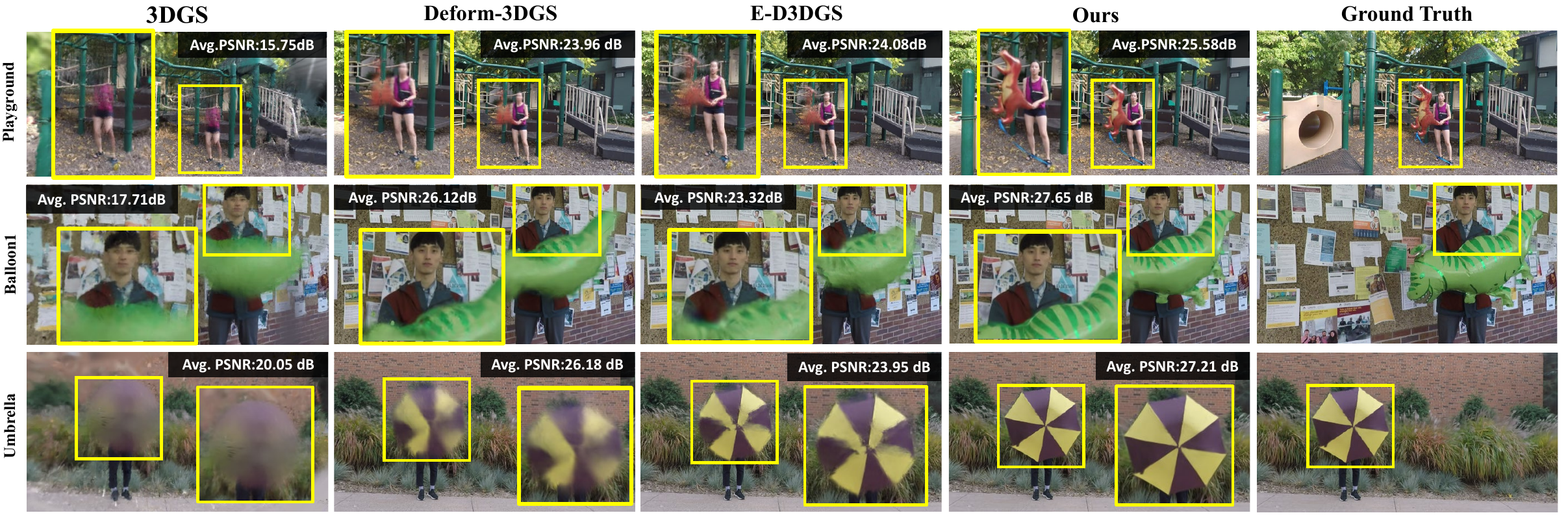}
    \caption{\textbf{Qualitative comparison on Nvidia Monocular dataset~\cite{gao2021dynamicviewsynthesisdynamic}.} Yellow boxes highlight zoomed-in regions for detail examination. Per-scene average PSNR values are provided.}
    \label{fig:nvidia-3scenes}
\end{figure*}
\begin{figure*}[t]
    \centering
    \includegraphics[width=1.0\textwidth]{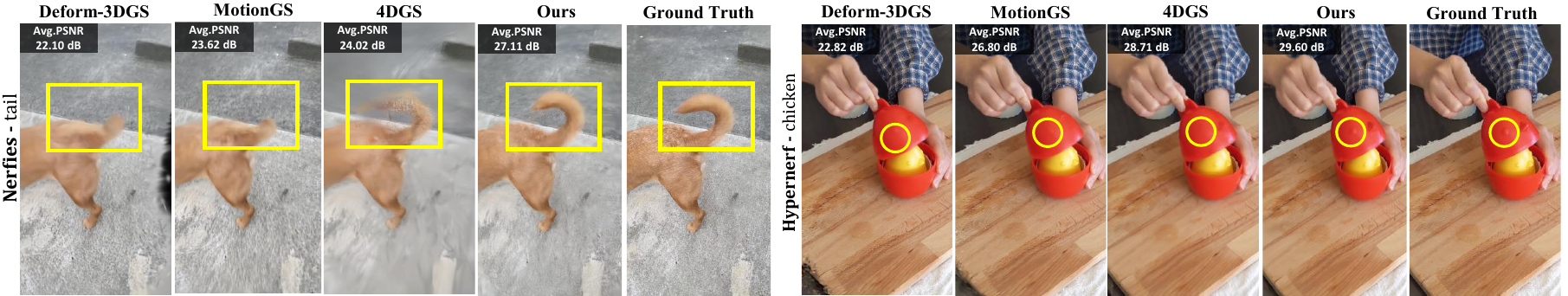}
    \caption{\textbf{Qualitative comparison on Nerfies-HyperNeRF dataset~\cite{park2021nerfies, park2021hypernerf}.} Yellow boxes highlight zoomed-in regions for detail examination. Per-scene average PSNR values are provided.}
    \label{fig:nerfies}
    \vspace{1em}
\end{figure*}

\section{Experiment}
\label{sec:Experiment}
\subsection{Experimental Setup}
\noindent \textbf{Implementation Details.} We implement our approach within a deformation-based dynamic scene rendering framework~\cite{wu20234d}, building the canonical space point cloud with a geometry foundation model~\cite{zhang2024monst3r}. Following~\cite{wu20234d}, we adopt a two-phase optimization strategy: a coarse phase where we freeze the deformation fields for dynamic regions while applying camera flow matching to static regions, followed by a fine phase that jointly optimizes both static and dynamic elements with active deformation fields. All experiments run on a single NVIDIA RTX A6000 GPU.

\noindent \textbf{Datasets and Metrics.} We evaluate our method on three representative dynamic scene datasets. The NVIDIA Monocular dataset~\cite{gao2021dynamicviewsynthesisdynamic} features forward-facing captures with controlled dynamics recorded by a camera array; following~\cite{kwak2025modec}, we use the provided monocular video sequences for benchmarking. Both Nerfies~\cite{park2021nerfies} and HyperNeRF~\cite{park2021hypernerf} datasets were captured using a vrig setup with synchronized stereo cameras, allowing us to train on one camera view while using the other for testing at identical timestamps. Performance is measured using PSNR, SSIM, and LPIPS~\cite{zhang2018unreasonable}, with metrics computed per frame and averaged across test sequences.
\subsection{Experimental Results}
\noindent \textbf{Quantitative Comparison.} We present quantitative results in Tables \ref{tab:nvidia-monocular} and \ref{tab:nerfies-hypernerf}. Our method consistently outperforms existing approaches across all eight scenes in the NVIDIA Monocular dataset on PSNR, SSIM, and LPIPS metrics, particularly in Balloon2, Playground, and Umbrella scenes, where we improve PSNR by over 2dB compared to recent methods like MoDec-GS~\cite{kwak2025modec} that employ complex motion modeling. On the Nerfies-HyperNeRF dataset, our approach also achieves best performance, surpassing all competing methods by significant margins.

\begin{table}[t]
\centering
\caption{\textbf{Quantitative comparison on Nerfies-HyperNeRF dataset~\cite{park2021nerfies, park2021hypernerf}.} We highlight the \best{best} and \second{second best} results.}
\label{tab:nerfies-hypernerf}
\small
\setlength{\tabcolsep}{6pt}
\scalebox{1.0}{\begin{tabular}{l|cc}
\toprule
\multirow{2}{*}{Methods} & \multicolumn{2}{c}{(a) Nerfies-HyperNeRF} \\
\cmidrule(lr){2-3}
 & PSNR$\uparrow$ & SSIM$\uparrow$ \\
\midrule
T-NeRF~\cite{pumarola2021d} & 20.43 & 0.508 \\
NSFF~\cite{li2021neural} & 19.41 & 0.423 \\
Nerfies~\cite{park2021nerfies} & 20.08 & 0.462 \\
HyperNeRF~\cite{park2021hypernerf} & 20.40 & 0.467 \\
Deformable-3DGS~\cite{yang2023deformable} & 22.19 & 0.558 \\
4DGS~\cite{wu20234d} & \second{24.46} & \second{0.588} \\
Ours & \best{25.97} & \best{0.674} \\
\midrule
\multirow{2}{*}{Methods} & \multicolumn{2}{c}{(b) HyperNeRF} \\
\cmidrule(lr){2-3}
 & PSNR$\uparrow$ & SSIM$\uparrow$ \\
\midrule
T-NeRF~\cite{pumarola2021d} & 21.31 & 0.583 \\
NSFF~\cite{li2021neural} & 19.02 & 0.456 \\
Nerfies~\cite{park2021nerfies} & 20.22 & 0.535 \\
HyperNeRF~\cite{park2021hypernerf} & 20.73 & 0.544 \\
Deformable-3DGS~\cite{yang2023deformable} & 22.48 & 0.611 \\
SC-GS~\cite{huang2024sc} & 21.20 & 0.576 \\
4DGS~\cite{wu20234d} & \second{25.16} & 0.684 \\
MotionGS~\cite{zhu2025motiongs} & 24.80 & \second{0.69} \\
MoDec-GS~\cite{kwak2025modec} & 25.02 & 0.679 \\
Ours & \best{26.15} & \best{0.736} \\
\bottomrule
\end{tabular}}
\vspace{1em}
\end{table}

\noindent \textbf{Qualitative Comparison.}
Figures \ref{fig:nvidia-3scenes} and \ref{fig:nerfies} show visual comparisons with existing methods. On the NVIDIA Monocular dataset, our method achieves better reconstruction quality, capturing the balloon and subtle ribbons in the Playground scene that other methods miss entirely, while maintaining sharp stripe patterns in Balloon1 that appear blurred in competing approaches.
On the Nerfies-HyperNeRF dataset, we successfully reconstruct the complete tail structure in the Tail sequence where alternatives produce incomplete results, and preserve the critical red exterior protrusion in the Chicken sequence that remains absent in other approaches. We also visualize the process of our self-correction flow matching in Figure~\ref{fig:image_warping} and its effects in Figure~\ref{fig:flow_matching_effects}, illustrating how the model captures subtle motions and aligns objects across frames.
More qualitative results are included in Appendix~\ref{sec:addtional_results}.

\subsection{Ablation Study} 
\label{subsec:ablation} 
\begin{figure}[t]
    \centering
    \includegraphics[width=\linewidth]{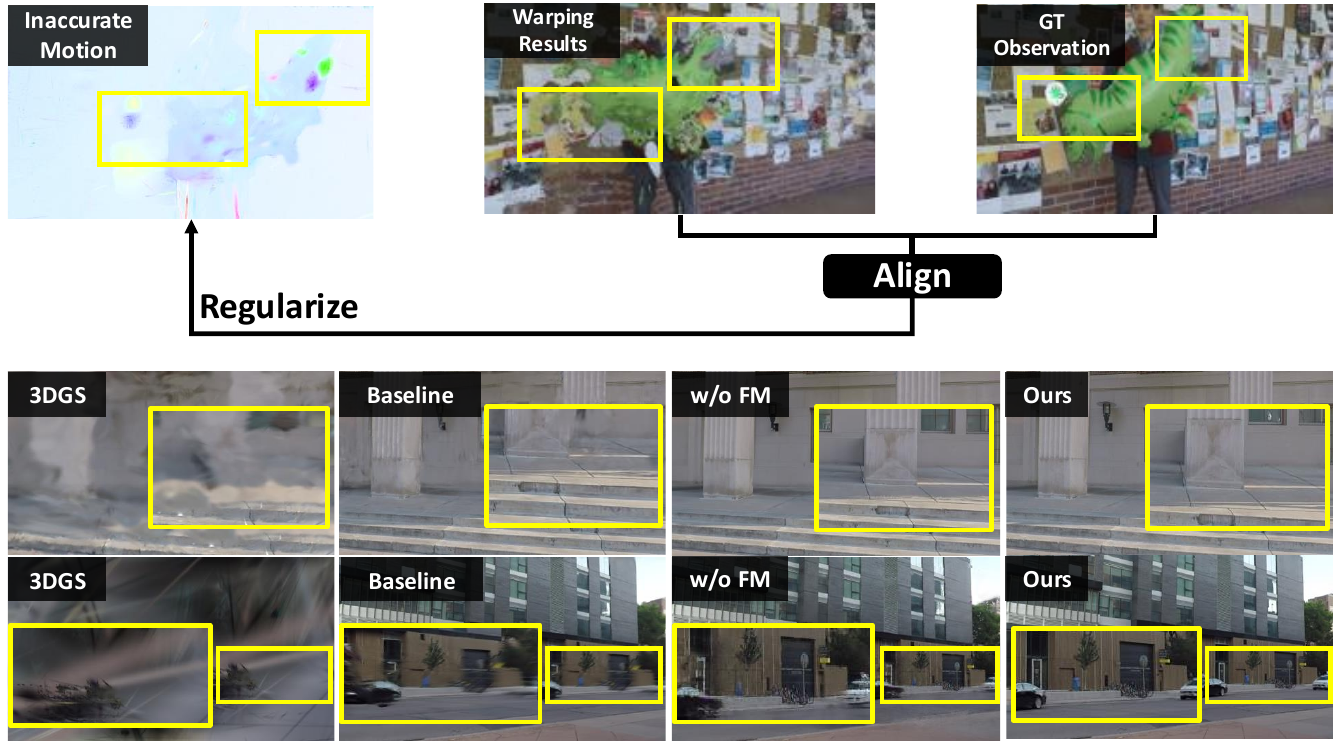}
    \caption{\textbf{Visualization of self-correction flow matching effects.}  
    \textbf{Top:} Incorrect motion leads to erroneous warping results that do not align with the ground truth, highlighted by yellow boxes, which in turn drives motion optimization.  
    \textbf{Bottom:} After flow matching, finer details are better reconstructed, demonstrating improved rendering quality.}
    \label{fig:flow_matching_effects}
    \vspace{-5pt}
\end{figure}

\begin{table}[t]
\centering
\caption{\textbf{Ablation study of key components in our method.} 
CompInit.: Complete Canonical Space Initialization; 
Sep.: Separation-based Dynamic Scene Modeling; 
FullFM: Full Flow Matching; 
CamFM: Camera Flow Matching. 
We highlight best performance in \textbf{bold}.}
\label{table:ablation_study}
\footnotesize
\setlength{\tabcolsep}{2.5pt}
\begin{tabular}{c|cccc|ccc}
\toprule
\# & CompInit. & Sep. & FullFM & CamFM & PSNR$\uparrow$ & SSIM$\uparrow$ & LPIPS$\downarrow$ \\
\midrule
Baseline &     &     &     &     & 25.81 & 0.844 & 0.210 \\
1 &     &     & \checkmark &  & 26.45 & 0.865 & 0.172 \\
\midrule
2 &  w/o align   &     &     &     & 24.50 & 0.815 & 0.260 \\
3 & \checkmark &     &     &     & 26.60 & 0.870 & 0.167 \\
4 & \checkmark & \checkmark &     &     & 27.00 & 0.880 & 0.159 \\
5 & \checkmark & \checkmark & \checkmark &     & 27.85 & 0.897 & 0.116 \\
Ours & \checkmark & \checkmark & \checkmark & \checkmark & \textbf{28.20} & \textbf{0.903} & \textbf{0.103} \\
\bottomrule
\end{tabular}
\end{table}

We analyze each component's effectiveness through a systematic ablation study, summarized in Table~\ref{table:ablation_study}. All experiments are conducted on the NVIDIA Monocular dataset.

\noindent \textbf{Effectiveness of flow matching.}
\textit{\textbf{Configuration \#1}} applies our self-correction flow matching directly on the baseline model, already yielding a notable improvement to 26.45dB PSNR (+0.64dB). This highlights the immediate effectiveness of our approach—without requiring additional motion guidance or architectural modifications, the model benefits from motion-aware supervision derived solely from 2D video cues even in a sparse initialization setup. As illustrated in Fig.~\ref{fig:flow_matching_effects}, our self-correction flow matching clearly correct the process of motion learning using only the 2D motion cues from input video.

\noindent \textbf{Impact of initialization and modeling.}
We investigate the effectiveness of our self-correction flow matching framework in building a complete and disentangled scene representation. Starting from the baseline, naively incorporating raw, unaligned point clouds (\textit{\textbf{configuration \#2}}) leads to significant degradation (\textminus0.71dB), indicating that misaligned geometric priors introduce noise rather than providing useful guidance, as also reflected in increased LPIPS scores. In contrast, our hierarchical alignment strategy (\textit{\textbf{configuration \#3}}) constructs a coherent canonical space, improving synthesis quality to 26.60dB (+0.79dB). Building upon this, further disentangling the static and dynamic components of the scene (\textit{\textbf{configuration \#4}}) yields a notable performance boost (27.00dB, +1.19dB), demonstrating that motion-aware modeling enables more faithful and generalizable scene encoding. These results collectively highlight the importance of structured scene initialization and motion decomposition in enhancing representation quality under our self-correction paradigm.

\noindent \textbf{Flow matching thrives with appropriate foundation.} 
Combining hierarchical alignment with static-dynamic separation (\textit{\textbf{configuration \#4}}) establishes a robust foundation at 27.00dB. Upon this foundation, Full Flow Matching (\textit{\textbf{configuration \#5}}) delivers +0.85dB gain to reach 27.85dB. This confirms flow matching is most effective when applied to representations that distinguish between motion types, allowing the model to respect real-world motion precisely where and how it matters most. Finally, adding Camera Flow Matching (\textit{\textbf{Ours}}) further improves static region quality, resulting in an additional gain The quantitative gains of each step are summarized in Fig.~\ref{fig:image_warping} in Appendix, illustrating the progressive benefits of our design choices.

%% file: sec/5_conclusion.tex
\section{Conclusion}
\label{sec:conclusion}
We present ReFlow, a unified dynamic scene reconstruction framework that learns 3D motion in a self-correcting manner from monocular video.  
By exploiting the inherent relationship between 3D motion and 2D frame-to-frame differences, ReFlow introduces a simple yet effective flow matching mechanism to directly supervise scene dynamics from video.  
To enhance Gaussian initialization and support separate modeling of static and dynamic objects, ReFlow incorporates a Complete Canonical Space Construction module and a Separation-Based Dynamic Scene Modeling module.
These components enable ReFlow to effectively perform region-specific motion learning and scene reconstruction, with experiments demonstrating superior reconstruction quality in diverse dynamic scenarios, establishing a new self-correction paradigm for 4D reconstruction.

%% file: sec/X_suppl.tex

\clearpage
\setcounter{section}{0}
\maketitlesupplementary

\renewcommand{\thesection}{\Alph{section}}
\renewcommand{\thesubsection}{\thesection.\arabic{subsection}}
\renewcommand{\thesubsubsection}{\thesubsection.\arabic{subsubsection}}

This appendix complements the main paper by providing detailed implementation and optimization settings, as well as additional experimental results.
We also provide an \href{https://rosetta-leong.github.io/ReFlow_Page/}{project page} for more visualization. 

\section{Implementation Details}

\subsection{Complete Canonical Space Construction}
\label{init_explanation}

We construct the complete canonical space through a two-stage alignment process, leveraging a geometric foundation model. We employ the method of~\cite{zhang2024monst3r} as our base geometric foundation model, denoted as $\Phi$, due to its fine-tuning on dynamic scenes, which enables robust handling of complex scenarios with moving objects. For any input frame pair $(I_i, I_j)$, this model computes 3D point maps and their confidence:
\begin{equation}
(\mathbf{X}_{i}^{i}, \mathbf{X}_{i}^{j}, \mathbf{C}_{i}^{i}, \mathbf{C}_{i}^{j}) = \Phi(I_i, I_j)
\label{eq:foundation_model_output}
\end{equation}
where $\mathbf{X}_{i}^{i} \in \mathbb{R}^{H \times W \times 3}$ is the 3D point map of frame $i$ in $i$'s coordinate system, $\mathbf{X}_{i}^{j} \in \mathbb{R}^{H \times W \times 3}$ is the 3D point map of frame $j$ in $i$'s coordinate system, and $\mathbf{C}_{i}^{i}, \mathbf{C}_{i}^{j} \in \mathbb{R}^{H \times W}$ are their respective confidence maps. For the input video, we partition it into uniform clips of 25-30 frames, striking an optimal balance between sufficient temporal context and computational efficiency.

The overall alignment process minimizes a composite loss function $\mathcal{L}_{\text{align}}$. This loss integrates three essential components: (1) reprojection error minimization ensuring geometric consistency across frames, (2) camera trajectory smoothness regularization preventing discontinuous camera paths, and (3) dynamic constraints specifically designed to accommodate dynamic scene elements. The specific formulation of these components can be referred to in~\cite{zhang2024monst3r}. Our alignment proceeds in two stages:

\noindent\textbf{1. Coarse Keyframe Alignment.}
We first establish a globally consistent structure by aligning $K \ll N$ keyframes selected from the video. For each selected keyframe pair $(I_a, I_b)$, we compute their relative geometry using Equation~\ref{eq:foundation_model_output}. With these pairwise estimations, we construct a connectivity graph $G_K(V_K, E_K)$ where vertices $V_K$ represent keyframes and each edge $e \in E_K$ connects a keyframe pair. This reduced graph enables efficient global optimization of keyframe camera poses $K_{\text{pose}}$, intrinsics $K_{\text{intr}}$, and depth maps $K_{\text{depth}}$:
\begin{equation}
\min_{K_{\text{pose}}, K_{\text{intr}}, K_{\text{depth}}} \sum_{(a,b) \in E_K} \mathcal{L}_{\text{align}}(\mathbf{X}_a^a, \mathbf{X}_a^b, \mathbf{C}_a^a, \mathbf{C}_a^b)
\label{eq:coarse_align}
\end{equation}

\noindent\textbf{2. Fine-grained Intra-clip Alignment.}
With keyframes aligned, we proceed to refine the geometry of intermediate frames within each clip $S_k$. For temporally close frames $(I_i, I_j)$ within clip $S_k$ (where $|i-j| \leq \delta$, with $\delta$ being the clip length or a predefined window), we compute their relative geometry using Equation~\ref{eq:foundation_model_output}. We initialize the parameters for these frames using the results from the coarse keyframe alignment stage ($K_{\text{params}}$) and then optimize the poses $S_k^{\text{pose}}$, intrinsics $S_k^{\text{intr}}$, and depth maps $S_k^{\text{depth}}$ for all frames within the clip:
\begin{equation}
\min_{\{S_k^{\text{pose}}, S_k^{\text{intr}}, S_k^{\text{depth}}\}} \sum_{\mathclap{\substack{i,j \in S_k \\ |i-j| \leq \delta}}}
\mathcal{L}_{\text{align}}(\mathbf{X}_{i}^{i}, \mathbf{X}_{i}^{j}, \mathbf{C}_{i}^{i}, \mathbf{C}_{i}^{j}; K_{\text{params}})
\label{eq:fine_align}
\end{equation}
The loss $\mathcal{L}_{\text{align}}$ in this stage enforces local geometric consistency between neighboring frames. The term $K_{\text{params}}$ ensures global consistency by anchoring the optimization to the already aligned keyframes. 

In our practical implementations, we observed that the approach in~\cite{zhang2024monst3r}, even within the two-stage framework described above, can occasionally exhibit instability in complex scenes when applied without proper initialization. To address this issue, particularly for the Coarse Keyframe Alignment stage (Equation~\ref{eq:coarse_align}), we enhance the global graph optimization by incorporating two critical priors: camera parameters estimated by COLMAP~\cite{schonberger2016structure} and semantic motion masks derived from~\cite{ravi2024sam}. Notably, we maintain the COLMAP camera parameters fixed throughout the coarse optimization process, providing a stable reference frame for reconstruction. These initialization priors offer a practical aid in the reconstruction process, helping to guide the model toward more stable and consistent results.

\subsection{Separation-Based Dynamic Scene Modeling}
\label{model_explannation}
Here we provide a detailed mathematical formulation of our disentangled representation approach for static and dynamic scene components. Our method systematically separates scene elements based on their motion characteristics, with specialized representation schemes for each component.

For static scene components, we adapt a tri-plane-based 3D Gaussian representation using three orthogonal spatial feature planes \( F_s = \{F^{xy}, F^{xz}, F^{yz}\} \), where each feature plane \( F \in \mathbb{R}^{H \times W \times C} \). 
Given a 3D point \((x,y,z)\), we query the Gaussian parameters by bilinearly interpolating each plane at its respective coordinate, concatenating the resulting features, and passing them through a static decoder \( D_s \):
\begin{equation}
\begin{aligned}
G_s(x,y,z) = D_s(&
   \text{Interp}(F^{xy}, x,y), \\
   & \text{Interp}(F^{xz}, x,z), \\
   & \text{Interp}(F^{yz}, y,z)
)
\end{aligned}
\end{equation}
Here, \( G_s = \{\mu_s, s_s, q_s, \sigma_s, c_s\} \) denotes the output Gaussian attributes, including position \( \mu_s \in \mathbb{R}^3 \), scale \( s_s \in \mathbb{R}_+^3 \), orientation quaternion \( q_s \in \mathbb{S}^3 \), opacity \( \sigma_s \in [0,1] \), and view-dependent appearance coefficients \( c_s \in \mathbb{R}^{D_c} \). These parameters remain time-invariant and represent the static part of the scene.

For dynamic scene components that change over time, we augment the static representation with temporal feature planes \( F_t = \{F^{xt}, F^{yt}, F^{zt}\} \) as described in Section~\ref{sec:modeling}, where each \( F^{it} \in \mathbb{R}^{D \times T \times C} \) encodes the interaction between spatial dimension \( i \in \{x,y,z\} \) and time \( t \).
For a 4D query \((x,y,z,t)\), we extract temporal features via interpolation along each spatial-time plane and concatenate them with the spatial features to form a comprehensive spatio-temporal feature:
\begin{equation}
\hat{F}_d(x,y,z,t) =
\text{Concat}_{F_{ij} \in \mathcal{F}_d} \text{Interp}(F_{ij}, i,j),
\end{equation}
where
\[
\mathcal{F}_d = \{ F^{xy}, F^{xz}, F^{yz}, F^{xt}, F^{yt}, F^{zt} \}.
\]
The decoder \(D_d\) then decodes this feature to produce the dynamic Gaussian parameters:
\begin{equation}
G_d(x,y,z,t) = D_d(\hat{F}_d(x,y,z,t)).
\end{equation}Our implementation of the dynamic Gaussians closely follows the approach in ~\cite{wu20234d}, using the same delta-based formulation for time-varying attributes. This ensures compatibility with established rendering pipelines while enabling our novel disentangled representation to effectively separate static and dynamic components.

\begin{figure}[t]
  \centering
  \includegraphics[width=1.0\linewidth]{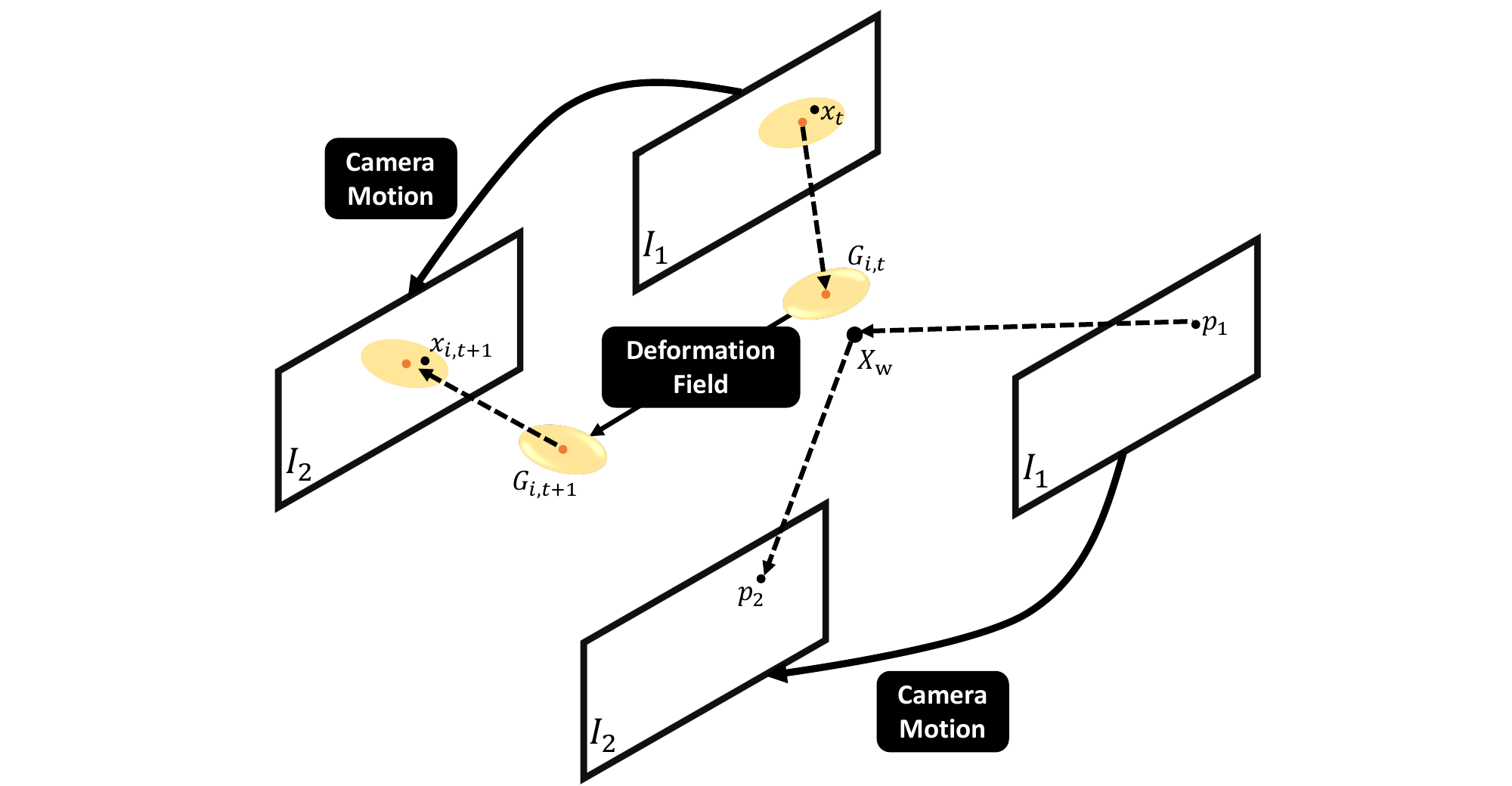}
  \caption{Illustration of the computation of Full Flow (top-left) and Camera Flow (bottom-right).}
  \label{fig:flow_calc}
\end{figure}

\subsection{Flow Render in Self-correction Flow Matching}
\label{sec:supp_flow_render_details}

In this section, we provide a more detailed exposition of the rendering processes for the \textit{Full Flow} ($\mathbf{F}_{full}$) and \textit{Camera Flow} ($\mathbf{F}_{cam}$), which are pivotal for our self-correction flow matching framework (see Fig.~\ref{fig:flow_calc}). 

\paragraph{Full Flow ($\mathbf{F}_{full}$)}
As described in \Cref{sec:flow_formulation}, the \textit{Full Flow} ($\mathbf{F}_{full}$) represents the overall pixel displacement caused by both object motion and camera movement, corresponding to changes across the entire image. Given the scene states $G_{t_1}$ and $G_{t_2}$ (e.g., 3D Gaussians at different timestamps) and the corresponding camera parameters $P_1$ and $P_2$ include both camera intrinsics and extrinsics, the Full Flow is synthesized via a rendering function:

\begin{equation}
\mathbf{F}_{full} = \text{FlowRender}(G_{t_1}, G_{t_2}, P_1, P_2).
\label{eq:supp_full_flow}
\end{equation}

Following~\cite{gao2024gaussianflow}, we formulate the $\mathbf{F}_{full}$ as a 2D motion field that represents the combined effects of camera movement and object movement in the scene. 
This can be interpreted as an optical flow field computed from 2D video frames, but crucially, it is derived from our 4D scene representation, which consists of 3D Gaussians, feature planes, and a deformation field.

Specifically, for a pixel $\mathbf{x}_t$ in the 2D image plane, we first normalize its position relative to each contributing $i$-th 2D Gaussian at time $t$. This transforms $\mathbf{x}_t$ into a \emph{canonical space}---a standard Gaussian distribution with zero mean and identity covariance, as referenced in \cite{gao2024gaussianflow}:
\begin{equation}
\hat{\mathbf{x}}_t = \mathbf{\Sigma}_{i,t}^{-1}(\mathbf{x}_t-\mathbf{\mu}_{i,t}).
\end{equation}
Here, $\mathbf{\mu}_{i,t}$ and $\mathbf{\Sigma}_{i,t}$ represent the center and covariance of the $i$-th 2D Gaussian, respectively, which are obtained by projecting its corresponding 3D Gaussian onto the camera plane at timestamp $t$. This projection inherently incorporates camera infomation.

Next, we transform this normalized pixel position into the future, assuming it moves with the $i$-th Gaussian. This is achieved by transforming $\hat{\mathbf{x}}_t$ back into the coordinate system of the $i$-th Gaussian at time $t+1$:
\begin{equation}
\mathbf{x}_{i, t+1} = \mathbf{\Sigma}_{i,t+1}\hat{\mathbf{x}}_t+\mathbf{\mu}_{i,t+1}.
\end{equation}

The parameters $\mathbf{\mu}_{i,t+1}$ and $\mathbf{\Sigma}_{i,t+1}$ denote the center and covariance matrix of the $i$-th Gaussian at timestamp $t+1$, which are derived by deforming the Gaussian from its state at time $t$ via a learned deformation field.
The individual flow contribution from the $i$-th Gaussian to the pixel $\mathbf{x}_t$ is then defined as the displacement between the original pixel position and its transformed future position:
\begin{equation}
\mathbf{F}_{full}^{i, t \to t+1} = \mathbf{x}_{i, t+1} - \mathbf{x}_{t}.
\end{equation}

Finally, the total flow for the pixel $\mathbf{x}_t$ is derived by accumulating the contributions from all relevant Gaussians in an $\alpha$-blending way:
\begin{align}
\mathbf{F}_{full} &= \sum_{i=1}^K w_i \mathbf{F}_{full}^{i, t \to t+1} \\
&= \sum_{i=1}^K w_i (\mathbf{x}_{i, t+1} - \mathbf{x}_{t}) ,
\end{align}
where $w_i$ is the weight of individual gaussians.

In practice, computing the Full Flow requires accounting for both the camera's egomotion and the motion of dynamic scene elements. To achieve this, we render the 2D Full Flow by projecting the 3D Gaussian deformations from time $t$ to $t+1$ according to the corresponding camera viewpoints $C_t$ and $C_{t+1}$. We adapt the implementation proposed by \cite{gao2024gaussianflow} for this process. This process ensures that the resulting 2D flow field accurately captures the total observed image-space displacement resulting from the combined motion of both the scene and the camera.

\paragraph{Camera Flow ($\mathbf{F}_{cam}$)}
As described in \Cref{sec:flow_formulation},
the \textit{Camera Flow} ($\mathbf{F}_{cam}$) isolates the 2D pixel displacements induced solely by the camera's egomotion, under the assumption that the 3D scene remains static. This flow component is particularly valuable as it provides a clean supervison signal in static scene regions, where observed pixel changes are attributable only to viewpoint shifts. The $\text{CamFlowRender}$ is employed to compute this flow:
\begin{equation}
    \mathbf{F}_{cam} = \text{CamFlowRender}(K_1, K_2, T_1, T_2, G_{t_1}),
    \label{eq:supp_cam_flow_def_main_ref_eng_direct}
\end{equation}
where $K_1$ and $K_2$ are the camera intrinsic matrices at times $t_1$ and $t_2$ respectively, $T_1$ and $T_2$ are the corresponding extrinsic matrices (i.e., camera poses), and $G_{t_1}$ denotes our 3D scene representation at time $t_1$ (e.g., a 3D Gaussian).

The computation of $\mathbf{F}_{cam}$ for a pixel $p_1$ in image $I_1$ (from camera $C_1$ at $t_1$) involves determining its new position $p_2$ in image $I_2$ (from camera $C_2$ at $t_2$), based on the premise that the underlying 3D scene point is stationary. Initially, $p_1$ (with homogeneous coordinates $\tilde{p}_1$) is back-projected to its 3D world coordinates $\mathbf{X}_w$. This step utilizes its depth $D_1(p_1)$ (rendered from $G_{t_1}$), the intrinsics $K_1$, and extrinsics $T_1$:
\begin{equation}
    \mathbf{X}_w = T_1^{-1} (K_1^{-1} (D_1(p_1) \tilde{p}_1)).
    \label{eq:supp_cam_flow_back_projection_eng_direct}
\end{equation}
Subsequently, this static 3D point $\mathbf{X}_w$ is projected into the image plane of camera $C_2$ (using $K_2$ and $T_2$) to yield the new 2D pixel location $p_2$:
\begin{equation}
    p_2 = \text{proj}(K_2 T_2 \mathbf{X}_w).
    \label{eq:supp_cam_flow_reprojection_eng_direct}
\end{equation}
The Camera Flow $\mathbf{F}_{cam}(p_1)$ is then defined as the displacement from $p_1$ to $p_2$:
\begin{equation}
    \mathbf{F}_{cam}(p_1) = p_2 - p_1.
    \label{eq:supp_cam_flow_final_calc_eng_direct}
\end{equation}
 The entire operation can be expressed compactly as:
\begin{equation}
\begin{split}
\mathbf{F}_{cam}(p_1) = \, & 
\text{proj}\Big( 
    K_2 T_2 \big( 
        T_1^{-1} \big( 
            K_1^{-1} \big( D_1(p_1) \tilde{p}_1 \big) 
        \big) 
    \big) 
\Big) \\
& - p_1
\end{split}
\label{eq:supp_cam_flow_summary_formula_eng_direct}
\end{equation}

In essence, the camera flow precisely characterizes the pixel-wise motion in static regions that arises solely from camera movement. By leveraging $\mathbf{F}_{cam}$ as a matching constraint, we ensure that static scene elements maintain consistent geometric relationships across time and viewpoints.

\subsection{Optimization Details}
\label{optimization appendix}

\paragraph{Optimization Strategy}
We adapt a two-stage optimization strategy following~\cite{yang2023deformable,wu20234d}. In the warm-up stage, deformation fields for dynamic regions are frozen and camera flow matching is applied only to static regions, stabilizing the static scene structure. In the subsequent full optimization stage, both static and dynamic Gaussians are jointly optimized. We adjust the weights of flow constraints according to scene complexity:
for scenes with simpler camera and object motion (like NVIDIA datasets), we use higher weights ($\lambda_{fullflow}=5.0$, $\lambda_{camflow}=0.3$) to enforce stronger constraints; for complex scenes with significant camera movement and object deformation (like HyperNeRF datasets) ($\lambda_{fullflow}=1.0$, $\lambda_{camflow}=0.1$), we use slightly lower weights to allow more flexibility.
For cross-time rendering, the weight is set to 0.1 times the motion consistency weight. All optimization is performed pairwise on video frames. Optimizer and other settings follow~\cite{wu20234d}.

\begin{table}[htbp]
  \centering
  \caption{\textbf{Training Time and Peak GPU Memory Usage.} We compare our method with the baseline on the Nvidia Monocular dataset.}
  \label{tab:training_comparison}
  \resizebox{\columnwidth}{!}{%
  \begin{tabular}{@{}l cccc@{}}
    \toprule
    \textbf{Method} & \textbf{Balloon1} & \textbf{Balloon2} & \textbf{dynamicFace} & \textbf{Jumping} \\
    \midrule
    Baseline~\cite{wu20234d} & 35m 32s / 6.30G & 37m 49s / 16.88G & 58m 23s / 6.24G & 37m 56s / 3.20G \\
    Ours            & 44m 52s / 7.53G & 46m 4s / 18.33G  & 68m 28s / 7.34G & 46m 41s / 4.53G \\
    \midrule
    \textbf{Method} & \textbf{Playground} & \textbf{Skating} & \textbf{Truck} & \textbf{Umbrella} \\
    \midrule
    Baseline~\cite{wu20234d} & 51m 19s / 6.45G & 37m 19s / 2.32G & 34m 56s / 4.12G & 44m 45s / 5.96G \\
    Ours            & 61m 9s / 7.50G & 47m 30s / 3.60G & 42m 58s / 5.61G & 54m 15s / 7.11G \\
    \bottomrule
  \end{tabular}%
  }
\end{table}

\begin{table}[htbp]
  \centering
  \caption{\textbf{Rendering Speed and Model Storage Size.}  We report the rendering frame rate (FPS) and the final model size (in MBs) for all scenes on the Nvidia Monocular dataset.}
  \label{tab:rendering_comparison}
  \resizebox{\columnwidth}{!}{%
  \begin{tabular}{@{}l cccc@{}}
    \toprule
    \textbf{Method} & \textbf{Balloon1} & \textbf{Balloon2} & \textbf{dynamicFace} & \textbf{Jumping} \\
    \midrule
    Baseline~\cite{wu20234d} & 20fps / 74M & 21fps / 72M & 15fps / 106M & 26fps / 35M \\
    Ours            & 18fps / 87M & 17fps / 86M & 13fps / 121M & 23fps / 52M \\
    \midrule
    \textbf{Method} & \textbf{Playground} & \textbf{Skating} & \textbf{Truck} & \textbf{Umbrella} \\
    \midrule
    Baseline~\cite{wu20234d} & 15fps / 93M & 30fps / 33M & 24fps / 45M & 25fps / 54M \\
    Ours            & 15fps / 105M & 25fps / 47M & 21fps / 65M & 21fps / 72M \\
    \bottomrule
  \end{tabular}%
  }
\end{table}

\begin{figure}[t]
    \centering
    \includegraphics[width=\linewidth]{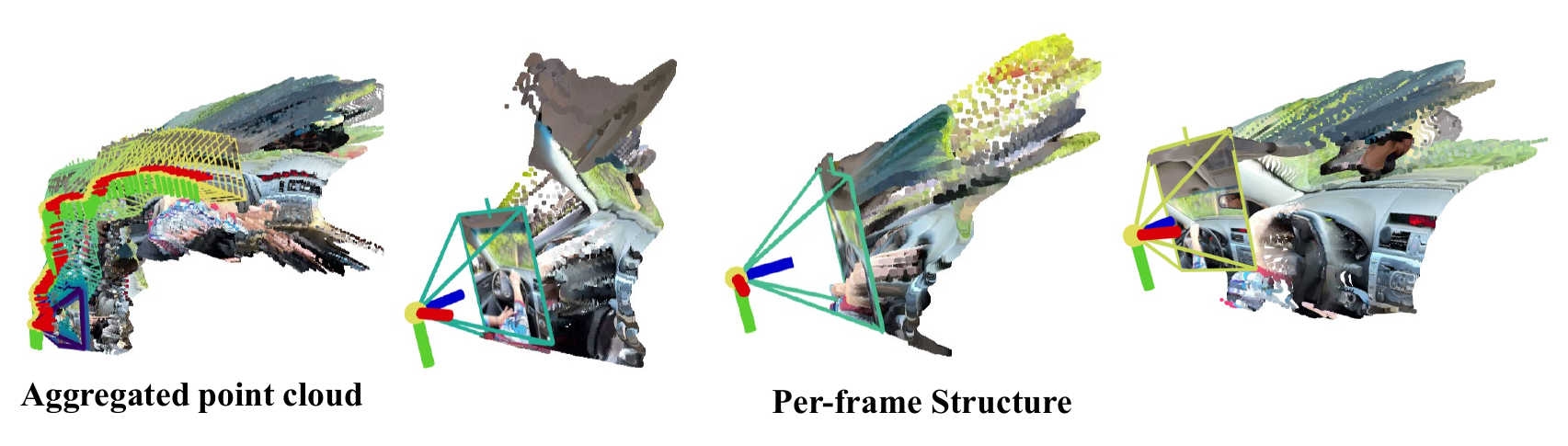}
    \caption{Failure case of the Utilized Geometry Foundation Model. \textbf{Top Left:} Aggregated point cloud exhibits noise and distortion from merging inconsistent frames. \textbf{Top Right:} Per-frame     structure reconstructions show significant noise and fragmentation, demonstrating its difficulty in establishing reliable geometry.}    
    \label{fig:point_cloud_limitations}
\end{figure}

\begin{figure*}[t]
    \centering
    \includegraphics[width=\linewidth]{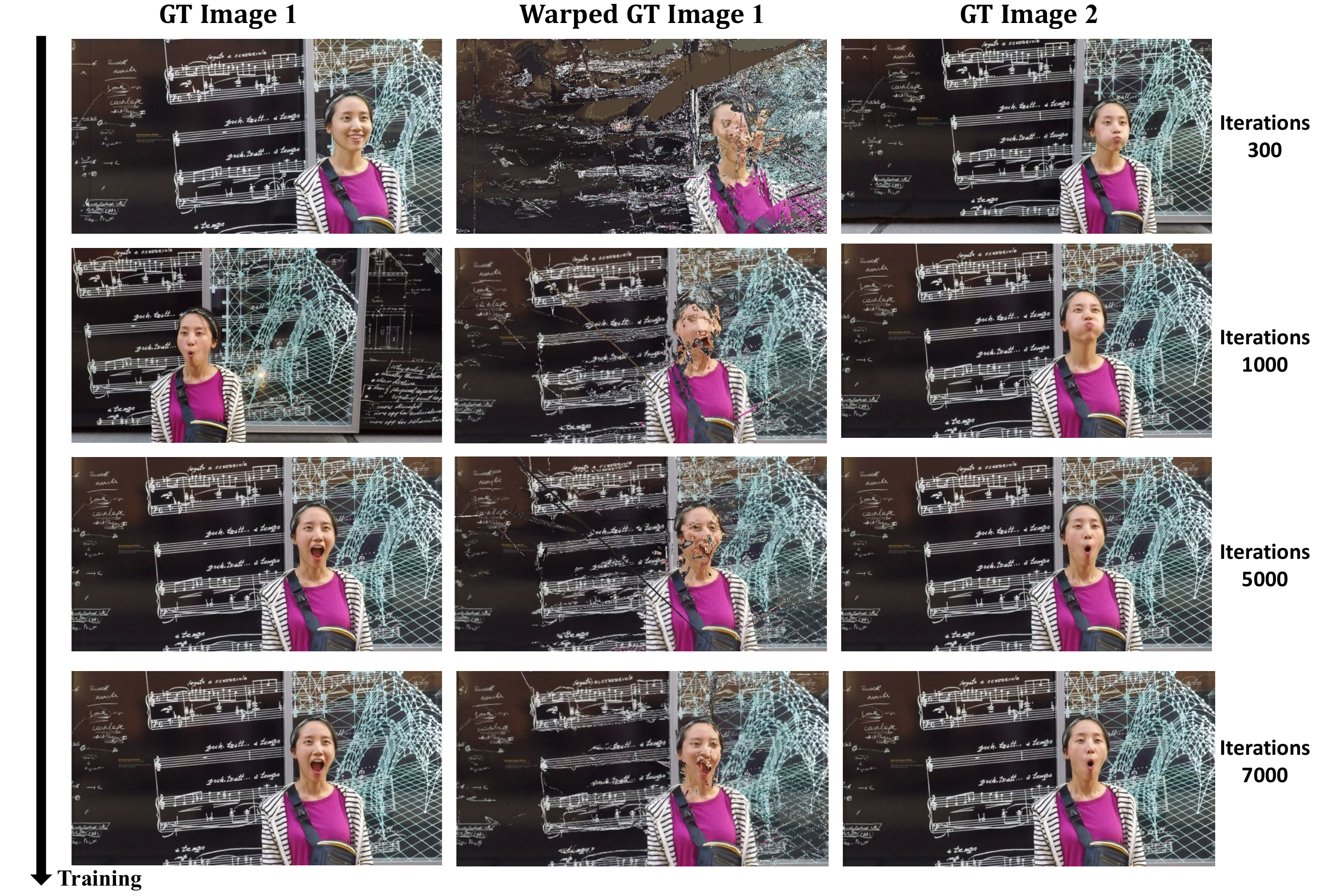}
    \caption{Visualization of our self-correction flow matching progress across training iterations, using the DynamicFace sequence from the Nvidia Monocular dataset~\cite{gao2021dynamicviewsynthesisdynamic}. Each row shows results at different training iterations: 300 (top row), 1000 (second row), 5000 (third row), and 7000 (bottom row). The columns present: \textbf{Left column ($I_1$):} Ground truth image at time $t_1$. \textbf{Middle column ($I_1^{warped}$):} Ground truth image from $t_1$ warped using our predicted flow field $\mathbf{F}$, demonstrating how content transforms to match the target frame. \textbf{Right column ($I_2$):} Ground truth image at time $t_2$, serving as the reference for evaluating warping accuracy. As training progresses, $I_1^{warped}$ increasingly aligns with $I_2$, demonstrating that our self-correction flow matching effectively learns to model dynamic scene motion without requiring external motion guidances.}
    \label{fig:image_warping}
\end{figure*}

\paragraph{Compute Resources}
\label{appendix: compute}
All experiments reported in this paper were conducted using a single NVIDIA RTX A6000 GPU with 48GB of memory. Memory is primarily consumed by the geometry foundation model used during the canonical space construction stage of our method. Actual memory usage is  affected by the clip size used in our coarse-to-fine alignment strategy. Processing larger input clips consumes more memory because handling these larger windows directly dictates the memory overhead needed for the second-stage fine alignment. 

\paragraph{Runtime Analysis}
We report the per-scene training time, peak GPU memory usage, rendering speed, and storage for our method in Table~\ref{tab:training_comparison} and Table~\ref{tab:rendering_comparison}. Our approach introduces moderate increases in training time and memory due to additional flow rendering and canonical space construction.
The core of our method is on exploring a self-correction motion learning mechanism during training without adding extra computational burden during inference. Thus, the rendering speed and inference time of our method are comparable to the baseline 4DGS, ensuring improved reconstruction quality without sacrificing rendering efficiency.

\section{Dataset Details}
For the HyperNeRF~\cite{park2021hypernerf} and Nerfies~\cite{park2021nerfies} datasets, we adapt a resolution that is downsampled by a factor of 2 from the standard resolution, resulting in 536×960 pixels. For the NVIDIA~\cite{gao2021dynamicviewsynthesisdynamic} dataset, we implement the temporal interpolation setting as outlined in~\cite{kwak2025modec}. We sample one frame per timestamp sequentially from the multi-view video sequence, collecting a total of 192 monocular frames. To establish our test set, we systematically hold out every 8th frame from training, which gives us 168 frames for training and reserves 24 frames for testing. This protocol challenges our model to reconstruct 3D motion between captured timepoints not observed during training.

\section{Additional Results}
\label{sec:addtional_results}
\begin{figure*}[t]
    \centering
    \includegraphics[width=\textwidth]{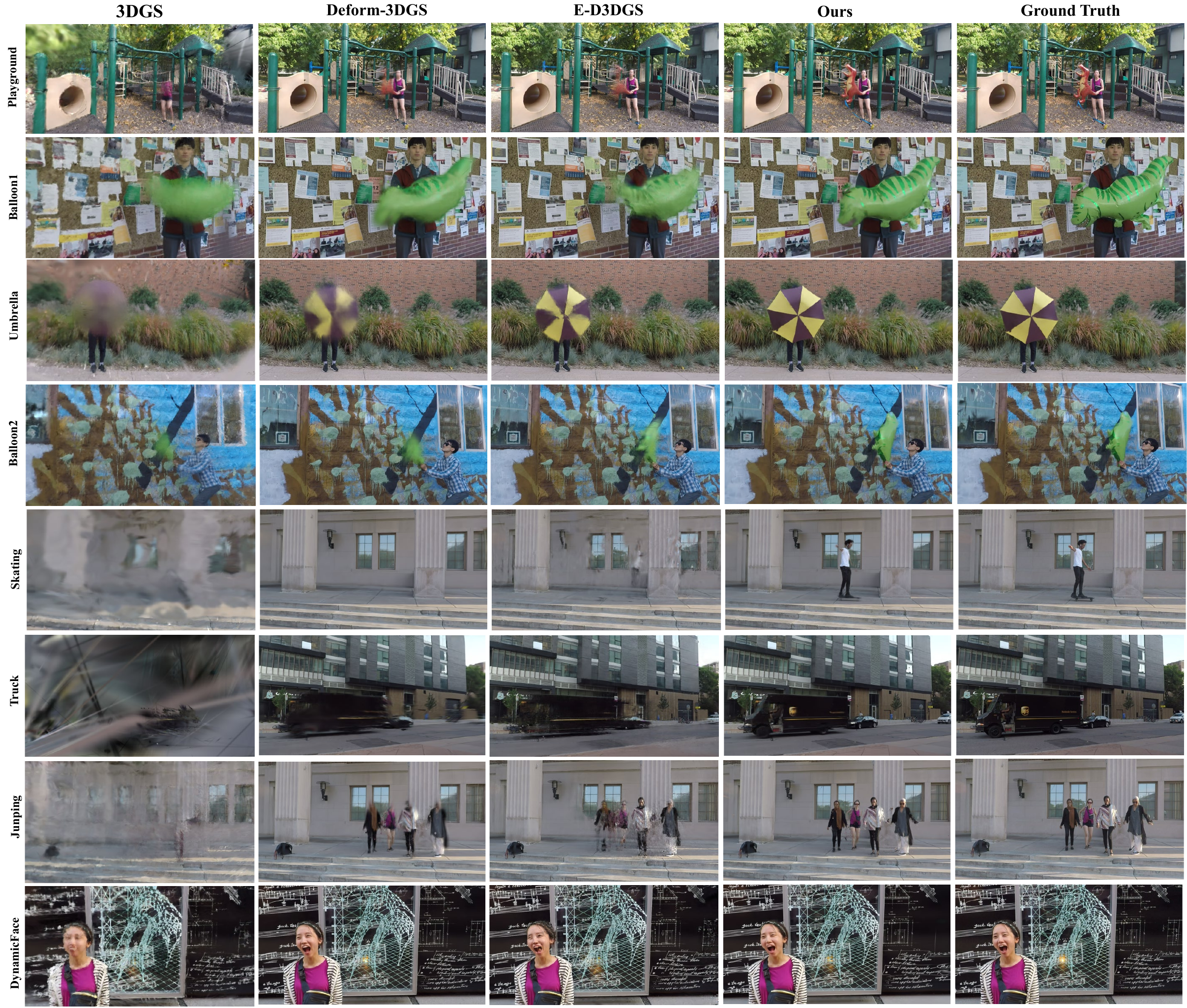}
    \caption{\textbf{Qualitative comparison on all scenes from the Nvidia Monocular dataset~\cite{gao2021dynamicviewsynthesisdynamic}.}}
    \label{fig:nvidia-all-scenes}
\end{figure*}

\begin{figure*}[t]
    \centering
    \includegraphics[width=\textwidth]{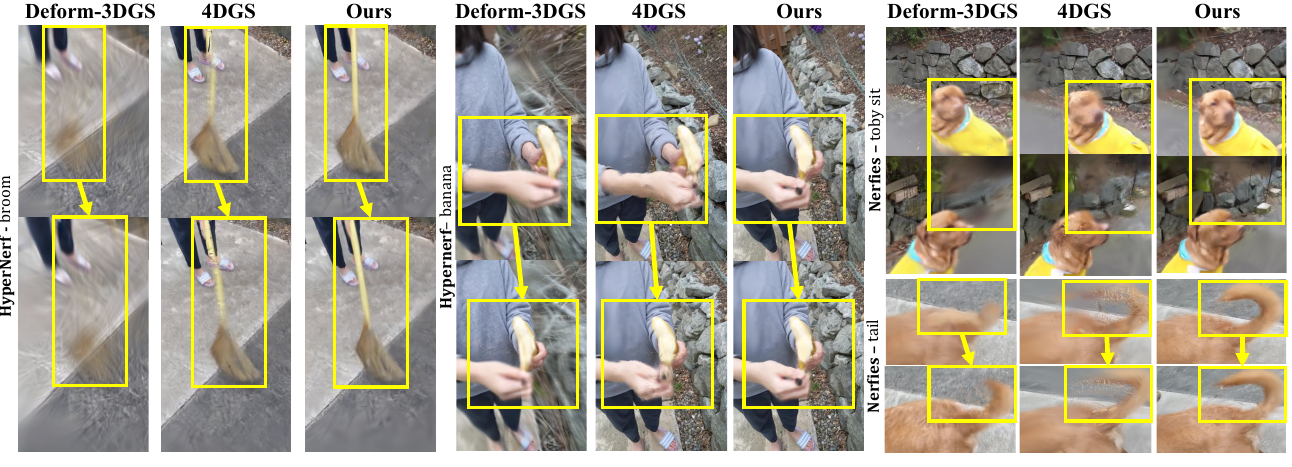}
    
    \caption{\textbf{More qualitative comparisons from the Nerfies-HyperNeRF dataset~\cite{park2021nerfies, park2021hypernerf}.}}
    \label{fig:hypernerf-all-scenes}
\end{figure*}

\begin{figure*}[t]
    \centering
    \includegraphics[width=\linewidth]{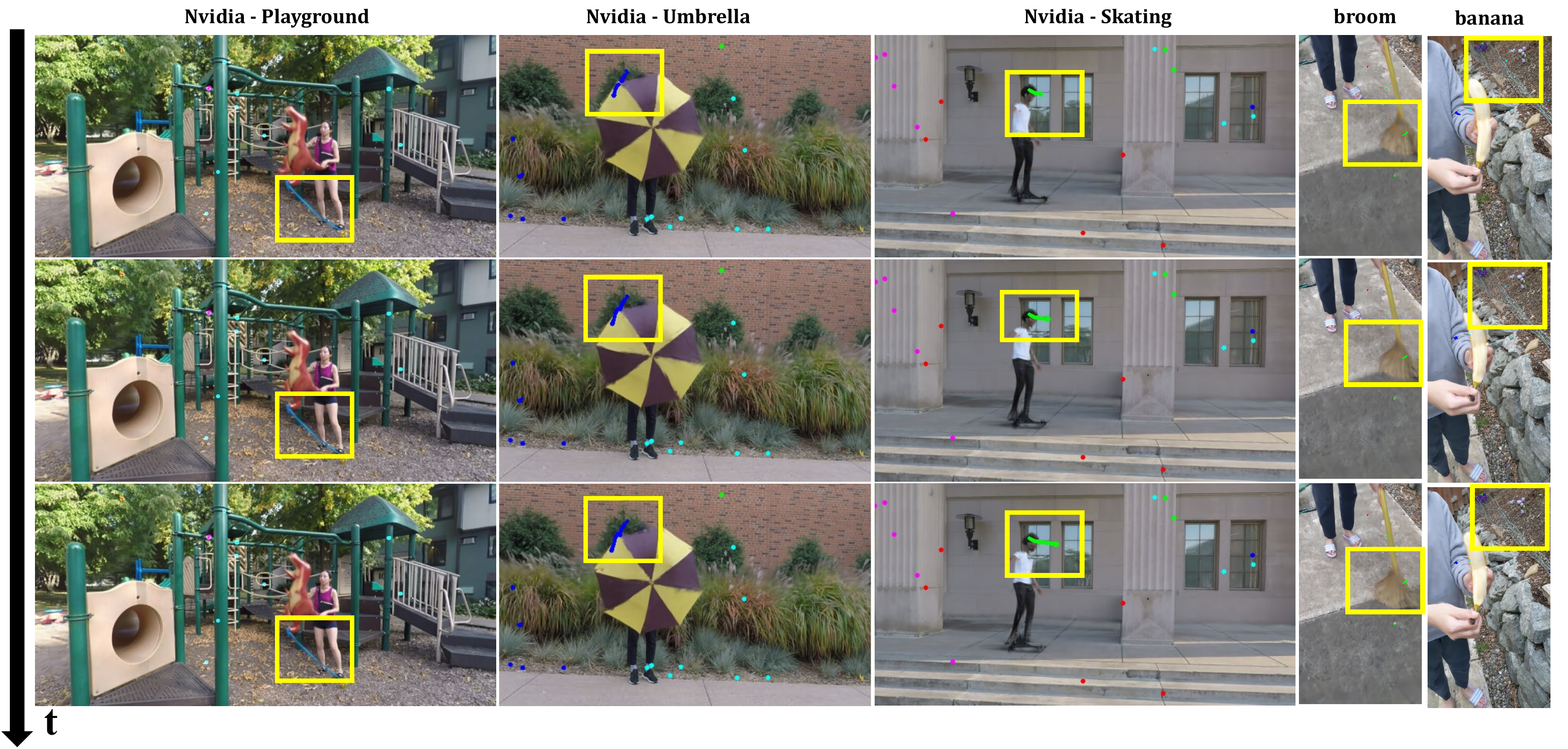}
    \caption{\textbf{Novel time synthesis results with trajectory visualization across different dynamic scenes.} Each column shows a different dataset: Nvidia-Playground, Nvidia-Umbrella, Nvidia-Skating, Hypernerf-broom, and Hypernerf-banana. The yellow boxes highlight dynamic/static regions with our tracked trajectories visualized using the Deform-GS approach. Rows represent different timesteps with fixed camera positions, demonstrating how our method correctly models temporal scene evolution. Note how static background elements remain perfectly stable across frames while dynamic components exhibit physically plausible motion paths. The Playground scene (leftmost column) particularly demonstrates our method's capability to preserve fine structures like blue ribbons during motion, which are typically challenging to reconstruct accurately.}
    \label{fig:novel_time_synthesis}
\end{figure*}

\paragraph{Detailed Per-scene Results} We visualize per-scene results for the NVIDIA dataset in \Cref{fig:nvidia-all-scenes}, and for the 
\begin{table}[htbp]
  \centering
  \caption{Quantitative comparison using PSNR($\uparrow$), SSIM($\uparrow$), and LPIPS($\downarrow$) metrics. Best results are in \textbf{bold}.}
  \label{tab:quantitative_comparison}
  \resizebox{\columnwidth}{!}{%
  \begin{tabular}{@{}l cccc@{}}
    \toprule
    \textbf{Method} & \textbf{Balloon1} & \textbf{Balloon2} & \textbf{Jumping} & \textbf{dynamicFace} \\
    \midrule
    \textbf{Ours} & \textbf{27.65} / \textbf{0.903} / \textbf{0.092} & \textbf{29.01} / \textbf{0.908} / \textbf{0.098} & 24.85 / \textbf{0.855} / \textbf{0.151} & 30.98 / 0.973 / 0.036 \\
    \textbf{Ours (External Flow)} & 27.35 / 0.899 / 0.095 & 27.74 / 0.843 / 0.179 & \textbf{24.92} / 0.853 / 0.162 & \textbf{31.10} / \textbf{0.975} / \textbf{0.034} \\
    \midrule[1.2pt] 
    \textbf{Method} & \textbf{Playground} & \textbf{Skating} & \textbf{Truck} & \textbf{Umbrella} \\
    \midrule
    \textbf{Ours} & \textbf{25.58} / \textbf{0.889} / \textbf{0.086} & \textbf{30.17} / 0.946 / 0.076 & \textbf{30.18} / \textbf{0.917} / \textbf{0.115} & 27.21 / \textbf{0.831} / \textbf{0.170} \\
    \textbf{Ours (External Flow)} & 20.97 / 0.820 / 0.099 & 30.09 / \textbf{0.976} / \textbf{0.032} & 29.30 / 0.905 / 0.133 & \textbf{27.35} / \textbf{0.831} / 0.171 \\
    \bottomrule
  \end{tabular}%
  }
\end{table}
Nerfies-HyperNeRF dataset in \Cref{fig:hypernerf-all-scenes}. The per-scene breakdown performance for Nerfies-HyperNeRF dataset is presented in \Cref{tab:nerfies-hypernerf perscene} and \Cref{tab:hypernerf perscene}. We also provide a visualization demonstrating our model's learned temporal dynamics in \Cref{fig:novel_time_synthesis}.

\paragraph{Comparison on External Flow Constraints.} 
We present a quantitative comparison between our self-correction approach and a variant that relies on external optical flow supervision (Ours (External Flow)) in~\Cref{tab:quantitative_comparison}. The key difference lies in the source of motion supervision. Approaches using external optical flow treat these estimates as pseudo-ground truth, forcing the learned 3D motion to follow potentially inaccurate 2D cues. This indirect supervision can conflict with the primary photometric reconstruction objective, particularly when the external flow contains errors. 
A notable example is the Playground scene, which features dominant camera motion and small, fast-moving foreground objects. External flow estimators primarily capture global camera-induced motion and often fail to represent fine-scale object dynamics, providing inaccurate motion supervision that misguides the learned 3D motion and results in a performance drop compared to our self-correction method.
Similar degradations are also observed in Balloon2 and Truck. 

In contrast, our key insight is that rather than aligning with indirect and often unreliable external priors, a more robust and principled approach is to derive motion supervision directly from the video itself. By grounding the learning signal directly ensure the reconstructed motion explains the observed frame-to-frame appearance changes, the learned motion naturally captures the true dynamics present in the video, rather than being forced to match external motion patterns. This approach embodies a more principled learning strategy for dynamic scene reconstruction and delivers improved robustness across diverse scenarios.


\section{Further Discussion}
\noindent\textbf{Limitation.} 
While our method learns 3D motion in a simple and self-correction way, the overall performance remains constrained by the quality of the scene geometry.
In scenes involving rapid object appearance/disappearance or topological changes, correspondence becomes unreliable, which fundamentally limits the geometry foundation model itself (see \Cref{fig:point_cloud_limitations}) and thus affects our entire pipeline. 
Nevertheless, the core contribution of our work lies in a novel self-correction motion supervision mechanism. This formulation is inherently flexible and can be readily integrated into a wide range of geometry foundation models or reconstruction pipelines. As geometry foundation models such as Align3r~\cite{lu2024align3r} and MegaSAM~\cite{li2024megasam} continue to advance, our framework can seamlessly benefit from improvements in these models for capturing more complex scenes.

\noindent\textbf{Discussion on Paradigms.} 
Recently, Feed-forward GS method and pretraining-based 4D reconstruction method have gained significant attention. To discuss these new paradigms alongside the per-scene optimization used in Reflow,
We provide a brief comparison in~\cref{tab:paradigm_comparison}. Pretraining-based methods like D4RT~\cite{zhang2025efficiently} achieves impressive results in 4D reconstruction, but only predict sparse point cloud and needs a large amount of annotated training data. Feed-forward GS pipelines like 4DGT~\cite{xu20254dgt} can be trained without geometry annotation, but still requires huge training resources and suffers from pixel-aligned GS representation. Per-scene optimization methods can be trained with limited resources, but are sensitive to initialization.
As a potential improvement, D4RT results can serve as the initialization of our methods, which can be further optimized and enable high-quality 4D reconstruction and real-time rendering.
\begin{table}[h]
\centering
\small
\vspace{-1em}
\caption{Comparison of 4D Reconstruction Paradigms.}
\label{tab:paradigm_comparison}
\vspace{-1em}
\begin{adjustbox}{width=0.8\linewidth}
\begin{tabularx}{1.6\linewidth}{@{}lXXX@{}} 
\toprule
 & \textbf{Pretraining-based} & \textbf{Feed-forward GS} & \textbf{Per-scene Optimization} \\
\midrule
\textbf{Methods} & D4RT~\cite{zhang2025efficiently} & DGS-LRM~\cite{lin2025dgs}, 4DGT~\cite{xu20254dgt} & Ours, 4DGS~\cite{wu20234d} \\
\midrule
\textbf{Representation} & Point Cloud & 3D Gaussian & 3D Gaussian \\
\midrule
\textbf{Annotation} & Video, Camera Pose, Point Cloud, Point Tracking  & Video, Camera Pose, External Priors (Depth/Flow) & Video, Camera Pose \\
\midrule
\textbf{Robustness} & Very High & High & Low \\ 
\midrule
\textbf{Resources} & 64 TPU & 64 GPU & Single GPU \\
\midrule
\textbf{Training Time} & 2 days & 15 days & 30 mins per scene \\
\midrule
\textbf{Limitation} & Sparse Rendering & Pixel-aligned GS & Sensitive to initialization \\
\bottomrule
\end{tabularx}
\end{adjustbox}
\vspace{-1em}
\end{table}

\noindent\textbf{Broader Impacts.} 
\label{appendix: social impacts}
This work poses no significant negative societal risks. 
Our framework enables self-correction 3D reconstruction and motion modeling from monocular video, making dynamic scene digitization more accessible. Users can convert everyday video footage—such as those captured by smartphones—into explicit 4D assets. These assets can facilitate downstream applications like content editing, digital scene creation, virtual reality, and educational visualization.
It focuses on general scenes and does not target personal identification or sensitive data collection.

\begin{table*}[t]
\centering
\caption{
\textbf{
Per-scene performance metrics on Nerfies-HyperNeRF~\cite{park2021nerfies, park2021hypernerf} dataset.} We highlight the \best{best} and \second{second best} results.}
\label{tab:nerfies-hypernerf perscene}
\small
\setlength{\tabcolsep}{4pt}
\begin{tabular}{l|cc|cc|cc}
\toprule
\multirow{2}{*}{Method} & \multicolumn{2}{c|}{Broom} & \multicolumn{2}{c|}{Tail} & \multicolumn{2}{c}{Toby-sit} \\
\cmidrule{2-7}
 & PSNR$\uparrow$ & SSIM$\uparrow$ & PSNR$\uparrow$ & SSIM$\uparrow$ & PSNR$\uparrow$ & SSIM$\uparrow$ \\
\midrule
T-NeRF~\cite{pumarola2021d} & 20.17 & 0.257 & 22.11 & 0.385 & 18.53 & 0.330 \\
NSFF~\cite{li2021neural} & 20.46 & 0.247 & 21.72 & 0.388 & 18.65 & 0.329 \\
Nerfies~\cite{park2021nerfies} & 19.51 & 0.202 & 21.17 & 0.305 & 18.41 & 0.326 \\
HyperNeRF~\cite{park2021hypernerf} & 19.23 & 0.197 & 21.13 & 0.301 & 18.33 & 0.324 \\
Deformable-3DGS~\cite{yang2023deformable} & 20.51 & 0.352 & 22.10 &  0.474 & 21.13 & 0.427 \\
4DGS~\cite{wu20234d} & \second{22.00} & \second{0.366} & \second{24.02} & \second{0.426} & \second{22.07} & \second{0.365} \\
Ours & \best{23.97} & \best{0.481} & \best{27.11} & \best{0.598} & \best{24.12} & \best{0.502} \\
\midrule
\multirow{2}{*}{Method} & \multicolumn{2}{c|}{3DPrinter} & \multicolumn{2}{c|}{Chicken} & \multicolumn{2}{c}{Peel-banana} \\
\cmidrule{2-7}
 & PSNR$\uparrow$ & SSIM$\uparrow$ & PSNR$\uparrow$ & SSIM$\uparrow$ & PSNR$\uparrow$ & SSIM$\uparrow$ \\
\midrule
T-NeRF~\cite{pumarola2021d} & 18.60 & 0.591 & 21.11 & 0.764 & 22.07 & 0.721 \\
NSFF~\cite{li2021neural} & 16.26 & 0.426 & 20.72 & 0.619 & 18.62 & 0.530 \\
Nerfies~\cite{park2021nerfies} & 18.81 & 0.588 & 22.71 & 0.742 & 19.85 & 0.609 \\
HyperNeRF~\cite{park2021hypernerf}  & 18.73 & 0.586 & 23.88 & 0.753 & 21.08 & 0.641 \\
Deformable-3DGS~\cite{yang2023deformable} & 20.53 & 0.641 & 22.82 & 0.618 & 26.05 & 0.833 \\
4DGS~\cite{wu20234d} & \second{21.99} & \second{0.704} & \second{28.65} & \second{0.814} & \second{28.01} & \second{0.852} \\
Ours & \best{22.91} & \best{0.738} & \best{29.60} & \best{0.868} & \best{28.13} & \best{0.858} \\
\bottomrule
\end{tabular}
\end{table*}
\begin{table*}[t]
\centering
\caption{
\textbf{
Per-scene performance metrics on HyperNeRF~\cite{park2021hypernerf} dataset.} We highlight the \best{best} and \second{second best} results.
}
\label{tab:hypernerf perscene}
\small
\setlength{\tabcolsep}{3.8pt}
\begin{tabular}{l|cc|cc|cc|cc}
\toprule
\multirow{2}{*}{Method} & \multicolumn{2}{c|}{3D Printer} & \multicolumn{2}{c|}{Chicken} & \multicolumn{2}{c|}{Broom} & \multicolumn{2}{c}{Banana} \\
\cmidrule{2-9}
 & PSNR$\uparrow$ & SSIM$\uparrow$ & PSNR$\uparrow$ & SSIM$\uparrow$ & PSNR$\uparrow$ & SSIM$\uparrow$ & PSNR$\uparrow$ & SSIM$\uparrow$ \\
\midrule
T-NeRF~\cite{pumarola2021d} & 18.60 & 0.591 & 24.41 & 0.764 & 20.17 & 0.257 & 22.07 & 0.721 \\
NSFF~\cite{li2021neural} & 16.26 & 0.426 & 20.72 & 0.619 & 20.46 & 0.247 & 18.62 & 0.530 \\
Nerfies~\cite{park2021nerfies} & 18.81 & 0.588 & 22.71 & 0.742 & 19.51 & 0.202 & 19.85 & 0.609 \\
HyperNeRF~\cite{park2021hypernerf} & 18.73 & 0.583 & 23.88 & 0.753 & 19.23 & 0.197 & 21.08 & 0.641 \\
Deformable-3DGS~\cite{yang2023deformable} & 20.53 & 0.641 & 22.82 & 0.618 & 20.51 & 0.352 & 26.05 & 0.833 \\
SC-GS~\cite{huang2024sc} & 18.79 &  0.613 &  21.85 & 0.616 & 18.66 & 0.269 &  25.49 & 0.806 \\
MoDec-GS~\cite{kwak2025modec} &  \second{22.00} & \second{0.706}  &  \second{28.77} & \second{0.834} &  \second{21.04} & \second{0.303} & \best{28.25} & \best{0.873} \\
Ours &  \best{22.91} & \best{0.738}  &  \best{29.60} & \best{0.868} & \best{23.97} & \best{0.481} & \second{28.13} & \second{0.858} \\
\bottomrule
\end{tabular}
\end{table*}

\section{Data and Code Availability}
\label{appendix: data}
The datasets used in this study are publicly available: NVIDIA Monocular~\cite{gao2021dynamic} (\url{https://github.com/gaochen315/DynamicNeRF}), Nerfies~\cite{park2021nerfies} (\url{https://github.com/google/nerfies/releases/tag/0.1}), and HyperNeRF~\cite{park2021hypernerf} (\url{https://github.com/google/hypernerf/releases/tag/v0.1}), all under permissive licenses. Our baseline code~\cite{wu20234d} is at \url{https://github.com/hustvl/4DGaussians}. 


\clearpage